\documentclass[journal]{IEEEtran}

\ifCLASSINFOpdf
\else
\fi

\usepackage{amssymb}

\usepackage{amsmath}

\usepackage{booktabs}

\usepackage{multirow}
\usepackage{color}
\usepackage[top=2cm, bottom=2cm, left=2cm, right=2cm]{geometry}
\usepackage[linesnumbered,ruled,lined]{algorithm2e}
\usepackage{algorithmicx}
\usepackage{algpseudocode}
\usepackage{amsmath}

\usepackage{diagbox}

\usepackage{threeparttable}

\usepackage{graphicx}
\usepackage{epstopdf}
\usepackage{subfigure}

\usepackage{CJK}

\usepackage{array}
\makeatletter
\newcommand{\thickhline}{%
    \noalign {\ifnum 0=`}\fi \hrule height 1pt
    \futurelet \reserved@a \@xhline
}
\newcolumntype{"}{@{\hskip\tabcolsep\vrule width 1pt\hskip\tabcolsep}}
\makeatother

\begin{document}

\title{Any Part of Bayesian Network Structure Learning}

\author{Zhaolong~Ling,
        Kui~Yu,
        Hao~Wang,
        Lin Liu,
        and~Jiuyong Li

\IEEEcompsocitemizethanks{\IEEEcompsocthanksitem This work is partly supported by the National Key Research and Development Program of China (under grant 2019YFB1704101), and the National Science Foundation of China (under grant 61876206 and 61872002). 
\IEEEcompsocthanksitem Z. Ling is with the School of Computer Science and Technology, Anhui University, Hefei, Anhui, 230601, China.
E-mail: zlling@ahu.edu.cn.
\IEEEcompsocthanksitem K. Yu and H. Wang are with Key Laboratory of Knowledge Engineering with Big Data of Ministry of Education (Hefei University of Technology), and the School of Computer and Information, Hefei University of Technology, Hefei, Anhui, 230009, China.
E-mail: yukui@hfut.edu.cn, jsjxwangh@hfut.edu.cn.
\IEEEcompsocthanksitem L. Liu and J. Li are with the School of Information Technology and Mathematical Sciences, University of South Australia, Adelaide, SA, 5095, Australia. E-mail: lin.liu@unisa.edu.au, jiuyong.li@unisa.edu.au.}
}

\markboth{IEEE TRANSACTIONS ON CYBERNETICS,~Vol.~14, No.~8, August~2020}%
{Shell \MakeLowercase{\textit{et al.}}: Bare Demo of IEEEtran.cls for IEEE Journals}

\maketitle

\begin{abstract}

We study an interesting and challenging problem, learning any part of a Bayesian network (BN) structure. 
In this challenge, it will be computationally inefficient using existing global BN structure learning algorithms to find an entire BN structure to achieve the part of a BN structure in which we are interested. And local BN structure learning algorithms encounter the false edge orientation problem when they are directly used to tackle this challenging problem.
In this paper, we first present a new concept of Expand-Backtracking to explain why local BN structure learning methods have the false edge orientation problem, then propose APSL, an efficient and accurate \underline{A}ny \underline{P}art of BN \underline{S}tructure \underline{L}earning algorithm. Specifically, APSL divides the V-structures in a Markov blanket (MB) into two types: collider V-structure and non-collider V-structure, then it starts from a node of interest and recursively finds both collider V-structures and non-collider V-structures in the found MBs, until the part of a BN structure in which we are interested are oriented. 
To improve the efficiency of APSL, we further design the APSL-FS algorithm using \underline{F}eature \underline{S}election, APSL-FS.
Using six benchmark BNs, the extensive experiments have validated the efficiency and accuracy of our methods.

\end{abstract}

\begin{IEEEkeywords}
Bayesian network, Local structure learning, Global structure learning, Feature selection.
\end{IEEEkeywords}

\IEEEpeerreviewmaketitle

\section{Introduction}

\IEEEPARstart{B}{ayesian} networks (BNs) are graphical models for representing multivariate probability distributions~\cite{pearl1988morgan,cooper1997simple,guyon2007causal}. The structure of a BN takes the form of a directed acyclic graph (DAG) that captures the probabilistic relationships between variables. Learning a BN plays a vital part in various applications, such as classification~\cite{aliferis2010local2,tsamardinos2003towards}, feature selection~\cite{aliferis2010local1,ling2019bamb,wang2020towards}, and knowledge discovery~\cite{spirtes2000causation,yu2019MCFS}.

However, in the era of big data, a BN may easily have more than 1,000 nodes. For instance, $Munin$\footnote{http://www.bnlearn.com/bnrepository/discrete-massive.html\#munin4} is a well-known BN for diagnosis of neuromuscular disorders~\cite{andreassen1996evaluation}, which has four subnetworks, and three of them have more than 1,000 nodes. When we are only interested in one of subnetwork structures, if we can start from any one of nodes of this subnetwork and then gradually expands to learn only this subnetwork structure, it will be much more efficient than learning the entire BN structure.

Thus in this paper, we focus on learning any part of a BN structure, that is, learning  a part of a BN structure around any one node to any depth.
For example in Fig. 1, given a target variable, structure learning to a depth of 1 means to discover and distinguish the parents and children (PC) of the target variable, structure learning to a depth of 2 means to discover and distinguish the PC of each node in the target's PC on the basis of structure learning to a depth of 1, and so on.

Clearly, it is trivial to obtain any part of a BN structure if we can learn a global BN structure using a global BN structure learning algorithm~\cite{margaritis2000bayesian,tsamardinos2006max,pellet2008using}. However, learning a global BN structure is known as NP-complete~\cite{pearl1986fusion,chickering2004large}, and easily becomes non-tractable in large scale applications where thousands of attributes are involved~\cite{scanagatta2016learning,vidaurre2010learning}. Furthermore, it is not necessary and wasteful to find a global BN structure when we are only interested in a part of a BN structure.

\begin{figure}[t]
\centering
 \includegraphics[width=3.5in, height=1.95in]{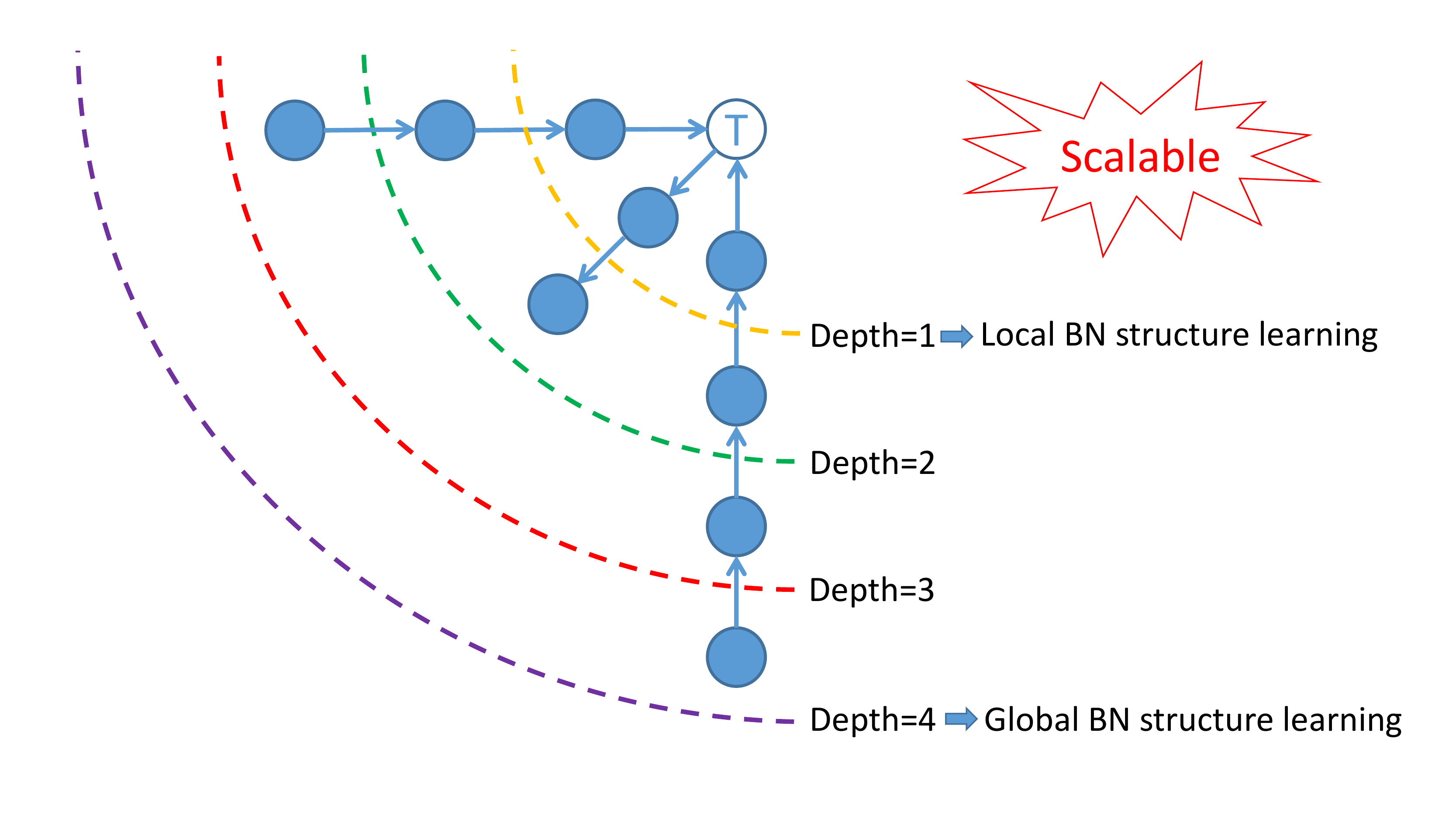}\vspace{-0.4cm}
 \caption{An illustrative example of learning a part of a BN structure around node $T$ to any depth from 1 to 4, which achieves a local BN structure around $T$ when learning to a depth of 1, and achieves a global BN structure when learning to a depth of 4 (the maximum depth).}
 \label{Figure 1}
\end{figure}

Recently, Gao et al.~\cite{gao2017local} proposed a new global BN structure learning algorithm, called Graph Growing Structure Learning (GGSL). Instead of finding the global structure directly,
GGSL starts from a target node and learns the local structure around the node using score-based local learning algorithm~\cite{gao2017socre}, then iteratively applies the local learning algorithm to the node's PC for gradually expanding the learned local BN structure until a global BN structure is achieved.
However, if we directly apply GGSL to tackle any part of BN structure learning problem, first, GGSL is still a global BN structure learning algorithm, and second, it is time-consuming or infeasible when the BN is large because the scored-based local learning algorithm~\cite{gao2017socre} used by GGSL needs to learn a BN structure involving all nodes selected currently at each iteration~\cite{ling2019bamb}.

Due to the limitation of the score-based local learning algorithm on large-sized BNs, existing local BN structure learning algorithms are constraint-based. Such as, PCD-by-PCD (PCD means Parents, Children and some Descendants)~\cite{yin2008partial} and Causal Markov Blanket (CMB)~\cite{gao2015local}. Local BN structure learning focus on discovering and distinguishing the parents and children of a target node~\cite{gao2015local}, and thus PCD-by-PCD and CMB only learn a part of a BN structure around any one node to a depth of 1. More specifically, both of PCD-by-PCD and CMB first find a local structure of a target node. If the parents and children of the target node cannot be distinguished in the local structure, these algorithms recursively find the local structure of the nodes in the target's PC for gradually expanding the learned local structure (Expanding phase), and then backtrack the edges in the learned expansive structure to distinguish the parents and children of the target (Backtracking phase). As illustrated in Fig. 2, we call this learning process Expand-Backtracking.

\begin{figure}[t]
\centering
 \includegraphics[width=1.9in, height=0.8in]{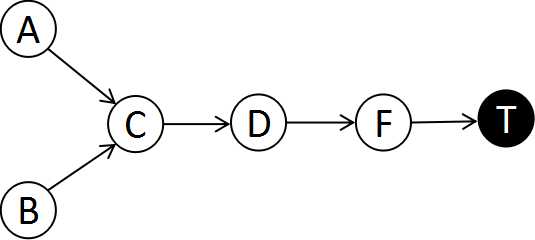}
 \caption{A simple Bayesian network. $T$ is a target node in black. Existing local BN structure learning algorithms cannot orient the edge $F- T$ when they only find the local structure of $T$. Then, they recursively find the local structure of the nodes $F, D$, and $C$ for expanding the local structure of $T$. Finally, since the V-structure $A\rightarrow C\leftarrow B$ can be oriented in the local structure of $C$, the local algorithms backtrack the edges $C \rightarrow D\rightarrow F\rightarrow T$, and thus $F$ is a parent of $T$.}
 \label{Figure 1}
\end{figure}

However, if we directly apply the local BN structure learning algorithms to tackle any part of BN structure learning problem,
this will lead to that many V-structures cannot be correctly found (i.e., V-structures missed) during the Expanding phase. Missing V-structures will generate many potential cascade errors in edge orientations during the Backtracking phase.

Moreover, PCD-by-PCD uses symmetry constraint (see Theorem 3 in Section III) to generate undirected edges, so it takes time to find more unnecessary PCs. CMB spends time tracking conditional independence changes after Markov blanket (MB, see Definition 6 in Section III) discovery, and the accuracy of CMB is inferior on small-sized data sets because it uses entire MB set as the conditioning set for tracking conditional independence changes.
Thus, even if the existing local BN structure learning algorithms do not miss the V-structures, they still cannot learn a part of a BN structure efficiently and accurately.


In this paper, we formally present any part of BN structure learning, to learn a part of a BN structure around any one node to any depth efficiently and accurately. As illustrated in Fig. 1, any part of BN structure learning can learn a local BN structure with a depth of 1, and achieve a global BN structure with a depth of the maximum depth. And hence, any part of BN structure learning has strong scalability. The main contributions of the paper are summarized as follows.

\begin{enumerate}


\item We present a new concept of Expand-Backtracking, to describe the learning process of the existing local BN structure learning algorithms. And we divide the V-structures included in an MB into collider V-structures and non-collider V-structures to analyze the missing V-structures in Expand-Backtracking.

\item Based on the analysis, we propose APSL, an efficient and accurate \underline{A}ny \underline{P}art of BN \underline{S}tructure \underline{L}earning algorithm.
    Specifically, APSL starts from any one node of interest and recursively finds both of the collider V-structures and non-collider V-structures in MBs, until all edges in the part of a BN structure are oriented.

\item We further design APSL-FS, an any part of BN structure learning algorithm using \underline{F}eature \underline{S}election. Specifically, APSL-FS employs feature selection for finding a local skeleton of a node without searching for conditioning sets to speed up local skeleton discovery, leading to improve the efficiency of APSL.

\item We conduct a series of experiments on six BN data sets, to validate the efficiency and accuracy of the proposed algorithms against 2 state-of-the-art local BN structure learning algorithms and 5 state-of-the-art global BN structure learning algorithms.

\end{enumerate}

The rest of this paper is organized as follows: Section II discusses related work. Section III provides notations and definitions. Section IV analyzes the missing V-structures in Expand-Backtracking. Section V presents the proposed algorithms APSL and APSL-FS. Section VI discusses the experimental results, and Section VII concludes the paper.


\section{Related Work}

Many algorithms for BN structure learning have been proposed and can be divided into two
main types: local methods and global methods. However, there are some issues with these methods when we apply them to tackle the any part of BN structure learning problem.

\textbf{Local BN structure learning algorithms} State-of-the-art local methods apply standard MB or PC discovery algorithms to recursively find V-structures in the local BN structure for edge orientations, until the parents and children of the target node are distinguished, and thus they learn a part of a BN structure around any one node to a depth of 1. PCD-by-PCD (PCD means Parents, Children and some Descendants)~\cite{yin2008partial} applies Max-Min Parents and Children (MMPC)~\cite{tsamardinos2003time} to recursively search for PC and separating sets, then uses them for local skeleton construction and finding V-structures, respectively, and finally uses the V-structures and Meek rules~\cite{meek1995causal} for edge orientations. However, at each iteration of any part of BN structure learning, since PCD-by-PCD only finds the V-structures connecting a node with its spouses V-structures, the V-structures included in the PC of the node are sometimes missed, then using the Meek-rules leads to false edge orientations in the part of a BN structure. Moreover, PCD-by-PCD uses symmetry constraint to generate undirected edges, so it needs to find the PC of each node in the target's PC to generate the undirected edges between the target and target's PC, which is time-consuming.
Causal Markov Blanket (CMB)~\cite{gao2015local} first uses HITON-MB~\cite{aliferis2003hiton} to find the MB of the target, then orients edges by tracking the conditional independence changes in MB of the target.
However, at each iteration of any part of a BN structure learning, since CMB only find V-structures included in the PC of a node, the V-structures connecting the node with its spouses are sometimes missed, then tracking conditional independence changes leads to false edge orientations in the part of a BN structure.
In addition, CMB uses entire MB set as the conditioning set and needs to spend time for conditional independence tests after MB discovery, which deteriorates the performance of CMB in accuracy and efficiency, respectively.

\textbf{Global BN structure learning algorithms} State-of-the-art global methods first identify each variable's MB/PC using the existing MB/PC methods, then construct a global BN skeleton (i.e., an undirected graph) using the found MBs/PCs, and finally orient the edge directions of the skeleton using constraint-based or score-based BN learning methods. Grow-Shrink (GS)~\cite{margaritis2000bayesian} first applies constraint-based MB method, Grow-Shrink Markov blanket (GSMB)~\cite{margaritis2000bayesian} to find MB of each node to construct global BN skeleton, then uses conditional independence test to find all V-structures, and finally orients undirect edges by using Meek-rules~\cite{meek1995causal}. Since then, many structure learning algorithms have been proposed. Max-Min Hill-Climbing (MMHC)~\cite{tsamardinos2006max} first applies constraint-based PC method, MMPC~\cite{tsamardinos2003time} to find PC of each node to construct global BN skeleton, then uses score-based method to orient edges. Both of Score-based Local Learning+Constraint (SLL+C)~\cite{niinimki2012local} and Score-based Local Learning+Greedy (SLL+G)~\cite{niinimki2012local} uses the score-based MB method, SLL~\cite{niinimki2012local} to find MB/PC of each node to construct global BN skeleton, then orient edges by using constraint-based and score-based methods, respectively. However, when we apply these global methods to any part of BN structure learning, it is time-consuming to learn an entire BN structure to achieve a part of a BN structure.

Recently, Gao et al.~\cite{gao2017local} proposed graph growing structure learning (GGSL) to learn a global BN structure. Instead of finding the MB/PC of each variable in advance, GGSL starts from any one node and learns the local structure around the node using the score-based MB discovery algorithm, S$^{2}$TMB~\cite{gao2017socre}, then iteratively applies S$^{2}$TMB to the node's neighbors for gradually expanding the learned local BN structure until an entire BN structure is achieved. However, GGSL still needs to learn an entire BN structure to achieve a part of a BN structure. In addition, although the score-based MB method can directly find the local BN structure without expanding outward, it is computationally expensive~\cite{ling2019bamb}, because it needs to learn a BN structure involving all nodes selected currently at each iteration. And hence, GGSL is time-consuming or infeasible when the size of a BN is large.

In summary, when we apply existing local and global BN structure learning algorithms to any part of BN structure learning, local methods are inaccurate and global methods are inefficient. Thus in this paper, we attempt to solve the problem of any part of BN structure learning.

\section{Notations and Definitions}

In the following, we will introduce the relevant definitions and theorems. Table I provides a summary of the notations used in this paper.

\textbf{Definition 1 (Conditional Independence)}~\cite{pearl2014probabilistic} Two variables $X$ and $Y$ are conditionally independent given $\textbf{Z}$, iff $P(X=x,Y=y|\textbf{Z}= z) = P(X=x|\textbf{Z}=z)P(Y=y|\textbf{Z} = z)$.

\textbf{Definition 2 (Bayesian Network)}~\cite{pearl2014probabilistic} Let $P$ be a discrete joint probability distribution of a set of random variables $\textbf{U}$ via a directed acyclic graph (DAG) $G$. We call the triplet $<\textbf{U}, G, P>$ a Bayesian Network (BN) if $<\textbf{U}, G, P>$ satisfies the \textbf{Markov Condition}: every variable in $\textbf{U}$ is conditionally independent of its non-descendant variables given its parents.

Markov condition enables us to recover a distribution $P$ from a known DAG $G$ in terms of conditional independence relationships. 

\begin{table}[t]
\centering
\caption{Summary of notations}
\footnotesize
\begin{tabular}{lll}
\toprule
Symbol&	Meaning \\
\midrule
$\textbf{U}$&                                        a variable set\\

$X,Y,T$&                                             a variable\\
$x,y$&		                                         a value of a variable\\

$Q$&                                                 a regular queue (first in, first out) \\
$\textbf{Z},\textbf{S}$&                             a conditioning set within $\textbf{U}$\\
$X\!\perp\!\!\!\perp Y|\textbf{Z}$ &                 $X$ is conditionally independent of $Y$ given $\textbf{Z}$\\
$X\not\!\perp\!\!\!\perp Y|\textbf{Z}$&              $X$ is conditionally dependent on $Y$ given $\textbf{Z}$\\

$\textbf{PC}_{T}$&                                   parents and children of $T$\\

$\textbf{SP}_{T}$&                                   spouses of $T$\\

$\textbf{SP}_{T}(X)$&                                   a subset of spouses of $T$, and each node in $\textbf{SP}_{T}(X)$\\
&  has a common child $X$ with $T$\\

$\textbf{V}$&                                        a queried variable set of variables\\
$\textbf{Sep}_{T}[X]$&                               a set that $d$-separates $X$ from $T$\\

$|.|$&                                               the size of a set\\

$SU(X;Y)$&           the correlation between $X$ and $Y$\\

\bottomrule
\end{tabular}

\label{tab:plain}
\end{table}

\textbf{Definition 3 (D-Separation)}~\cite{pearl2014probabilistic}. A path $p$ between $X$ and $Y$ given $\textbf{Z}\subseteq \textbf{U}\setminus\{X\cup Y\}$ is open, iff (1) every collider on $p$ is in $\textbf{Z}$ or has a descendant in $\textbf{Z}$, and (2) no other non-collider variables on $p$ are in $\textbf{Z}$. If the path $p$ is not open, then $p$ is blocked. Two variables $X$ and $Y$ are $d$-separated given $\textbf{Z}$, iff every path from $X$ to $Y$ is blocked by $\textbf{Z}$.

If two variables $X$ and $Y$ are d-separated relative to a set of variables $\textbf{Z}$ in a BN, such a set $\textbf{Z}$ would be called a separating set of $X$ from $Y$, then they are conditionally independent given $\textbf{Z}$ in all probability distributions where this BN can represent.

\textbf{Definition 4 (Faithfulness)}~\cite{spirtes2000causation}. A Bayesian network is presented by a DAG $G$ and a joint probability distribution $P$ over a variable set $\textbf{U}$.
$G$ is faithful to $P$ iff every conditional independence present in $P$ is entailed by $G$ and the Markov condition. $P$ is faithful iff there exists a DAG $G$ such that $G$ is faithful to $P$.

The faithfulness condition enables us to recover a DAG $G$ from a distribution $P$ to completely characterize $P$.

\textbf{Definition 5 (V-Structure)}~\cite{pearl2014probabilistic}. The triplet of variables $X$, $Y$, and $Z$ forms a V-structure if node $Z$ has two incoming edges from $X$ and $Y$, forming $X \rightarrow Z \leftarrow Y$, and $X$ is not adjacent to $Y$.

$Z$ is a collider if $Z$ has two incoming edges from $X$ and $Y$ in a path, respectively.

\textbf{Definition 6 (Markov Blanket)}~\cite{pearl2014probabilistic} Under the faithfulness assumption, given a target variable $T$, the Markov blanket of $T$ is unique and consists of parents, children, and spouses (other parents of the children) of $T$.

\textbf{Theorem 1}~\cite{spirtes2000causation} Under the faithfulness assumption, $X\in \textbf{U}$ and $ Y\in \textbf{U}$. If $X$ and $Y$ are adjacent, then $X \not\!\perp\!\!\!\perp Y|\textbf{S}$, $\forall \textbf{S}\subseteq \textbf{U}\setminus\{X\cup Y\}$.

\textbf{Theorem 2}~\cite{spirtes2000causation} Under the faithfulness assumption, $X\in \textbf{U}$, $Y\in \textbf{U}$, and $Z\in \textbf{U}$. If $X$, $Y$, and $Z$ forms the V-structure $X\rightarrow Z\leftarrow Y$, then $X\!\perp\!\!\!\perp Y|\textbf{S}$ and $X\not\!\perp\!\!\!\perp Y|\{\textbf{S}\cup Z\}$, $\forall \textbf{S}\subseteq \textbf{U}\setminus\{X\cup Y\cup Z\}$. $X$ is a spouse of $Y$.

Under the faithfulness assumption, Theorem 1 presents the property of PC, and Theorem 2 presents the property of spouses in an MB.

\textbf{Theorem 3} \emph{\textbf{Symmetry constraint.}}~\cite{gao2017efficient} Under the faithfulness assumption, if $X\in \textbf{PC}_{Y}$ exists, then $Y\in \textbf{PC}_{X}$ holds.


\section{Missing V-structures in Expand-Backtracking}


In this section, we first give the definition of Expand-Backtracking in Section IV-A, and then use two examples to analyze the missing V-structures in Expand-Backtracking in Section IV-B.

\subsection{Definition of Expand-Backtracking}

In this subsection, we first summarize the main ideas of local BN structure learning algorithms, then give the definition of the Expand-Backtracking.

Local BN structure learning aims to discover and distinguish the parents and children of a target variable, and thus the local BN structure learning algorithms are only able to learn a part of a BN structure around the target to a depth of 1. Moreover, existing local algorithms are constraint-based, because score-based local methods need to learn a BN structure involving all nodes selected currently at each iteration, which is time-consuming.

As constraint-based algorithms, local BN structure learning algorithms first find a local structure of a target node using the following three steps. Then, since the parents and children
of the target sometimes cannot be distinguished in the learned local structure, the local algorithms recursively apply these three steps to the target's neighbors for gradually expanding the learned local structure, until the parents and children of the target node are distinguished.

\begin{itemize}

\item[\emph{1)}] \emph{Skeleton identification.} Use standard local discovery algorithm to construct the local BN skeleton of a target node.
\item[\emph{2)}] \emph{V-structure discovery.} Discover V-structures in the local BN skeleton.
\item[\emph{3)}] \emph{Edge orientation.} Orient as many edges as possible given the V-structures in the learned part of BN skeleton, to get a part of BN structure around the target node.

\end{itemize}

Specifically, in the edge orientation step, given the discovered V-structures, local BN structure learning algorithms orient the edges not only in the local skeleton of a target node, but also the skeleton outside the local skeleton, to backtrack the edges into the parents and children of the target node for distinguishing them.

To facilitate the next step in presentation and analysis, we give the definition of the learning process of the local BN structure learning algorithms as follows.

\textbf{Definition 7 (Expand-Backtracking)} Under the faithfulness assumption, existing local BN structure learning algorithms first learn a local structure of a target node, then expand the learned local structure and backtrack the edges to distinguish parents and children of the target node. We call this learning process Expand-Backtracking.

Thus, V-structure discovery plays a crucial role in Expand-Backtracking. However, when the local BN structure learning algorithms are Expand-Backtracking, they ignore the correctness of the V-structures found (i.e., V-structures missed). Since the edge orientation step is based on the V-structure discovery step, missing V-structures in Expand-Backtracking will cause a cascade of false edge orientations in the obtained structure.




\subsection{Analysis of missing V-structures in Expand-Backtracking}

In this subsection, we first define two types of V-structures in an MB, then give the examples to demonstrate which type of V-structures cannot be correctly identified when the local BN structure learning algorithms are Expand-Backtracking.

\begin{figure}[t]
\centering
 \includegraphics[width=2.6in, height=1.9in]{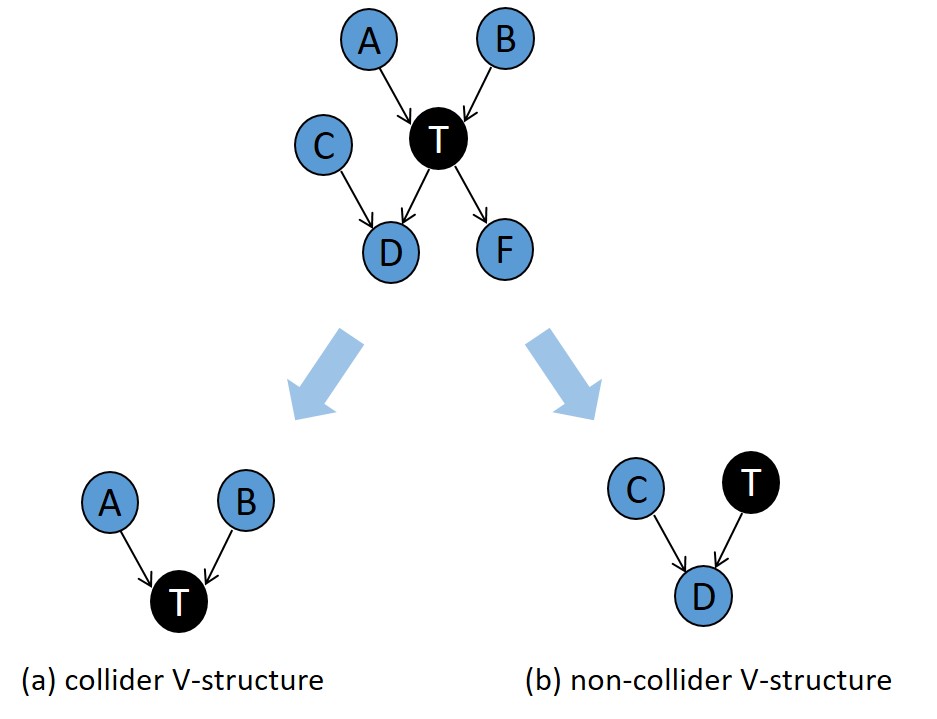}
 \caption{The Markov blanket (in blue) of node $T$ comprises $A$ and $B$ (parents), $D$ and $F$ (children), and $C$ (spouse). \protect\\  (a) Collider Vstructure ($T$ is a collider in the V-structure), and (b) non-collider V-structure ($T$ is not a collider in the V-structure).}
 \label{Figure 1}
\end{figure}

\begin{figure*}[t]
\centering
       \begin{tabular}{c|c|c}

      \subfigure{\includegraphics[width=2.2in, height=2.0in]{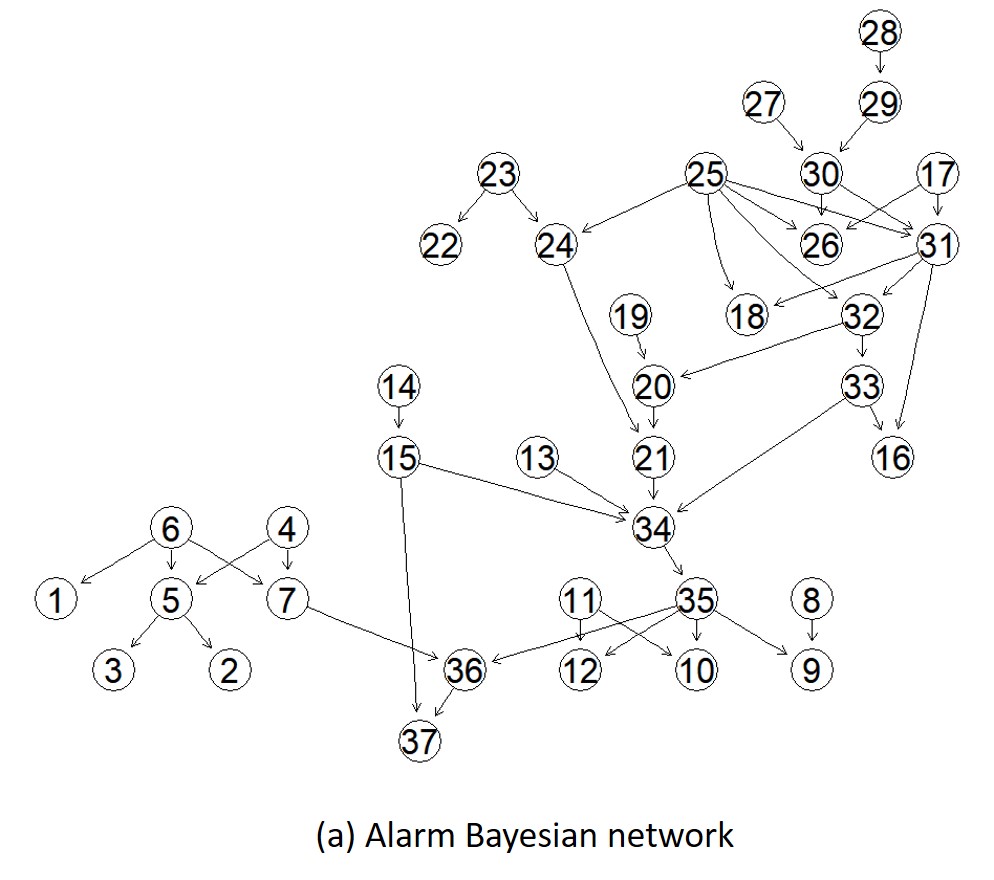}}&
      \subfigure{\includegraphics[width=2.2in, height=2.0in]{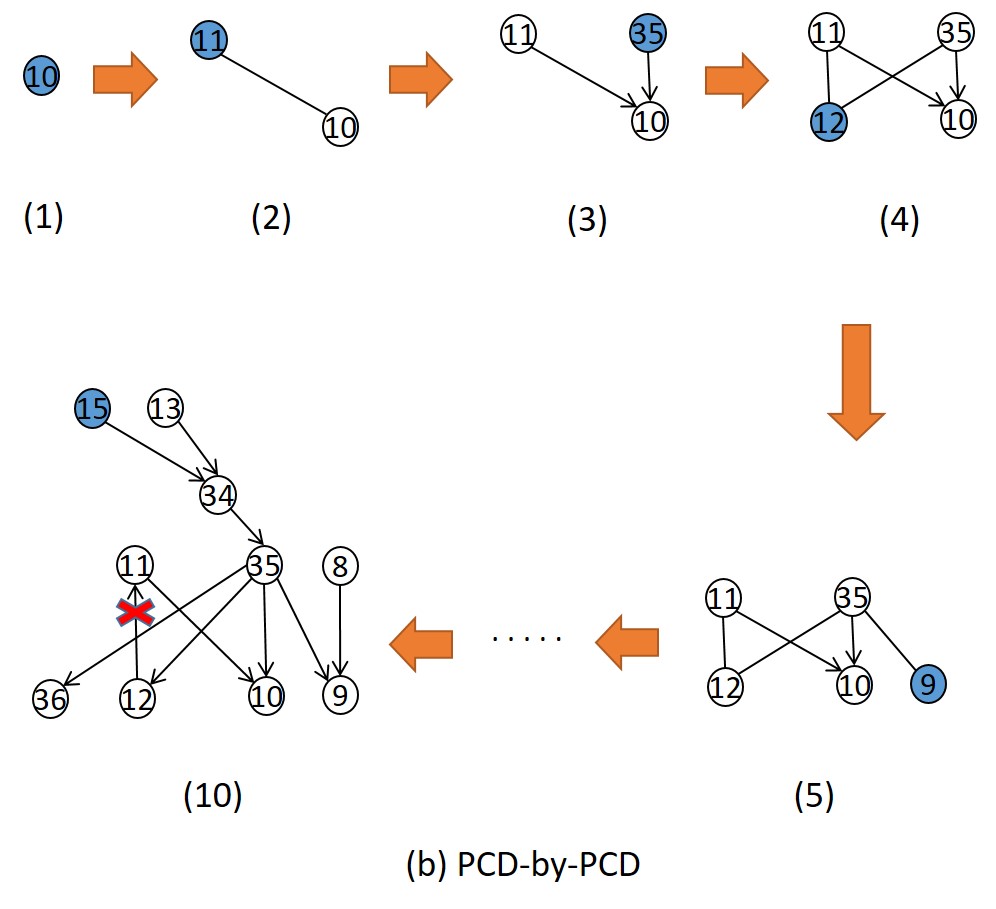}}&
      \subfigure{\includegraphics[width=2.15in, height=2.05in]{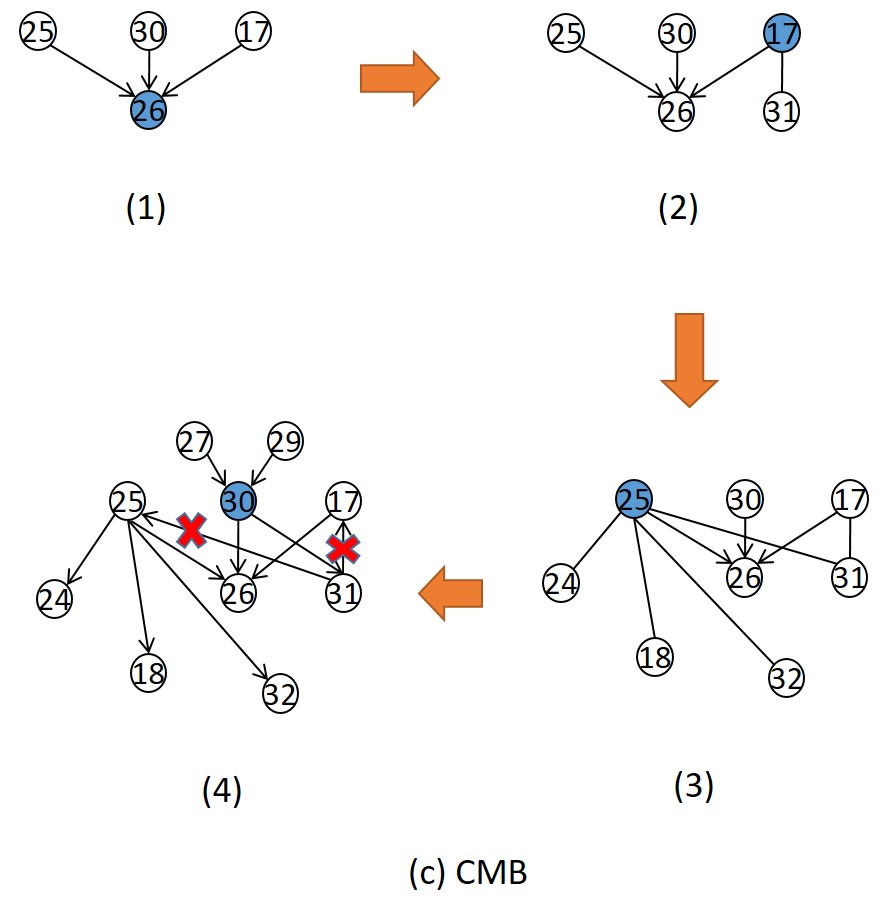}}\\

       \end{tabular}
 \caption{(a) The ALARM Bayesian network; (b) an example of using PCD-by-PCD to find a part of an Alarm Bayesian network structure around node 10 to a depth of 2; (c) an example of using CMB to find a part of an Alarm Bayesian network structure around node 26 to a depth of 2.
 \protect\\The red 'X' symbol denotes the falsely oriented edges, the blue node is the node that needs to find local structure at each iteration, the number in parentheses represents the level of iterations of an algorithm, and '$\cdots$' means omitted correctly oriented iterations.}
 \label{Figure 1}
\end{figure*}

\textbf{Definition 8 (Collider V-structure and Non-collider V-structure)} Under the faithfulness assumption, there are two types of the V-structure included in the MB of $T$, 1) collider V-structure: $T$ is a collider in the V-structure, and 2) non-collider V-structure: $T$ is not a collider in the V-structure.

Definition 8 gives two types of the V-structures included in an MB, as illustrated in Fig. 3. Thus, whether collider V-structures or non-collider V-structures cannot be correctly identified in the V-structure discovery step, it will cause the false edge orientations in the obtained structure. Below, we give the examples of the missing V-structures in Expand-Backtracking using two representative local BN structure learning algorithms.

\textbf{1) Missing collider V-structures}: PCD-by-PCD~\cite{yin2008partial} is a state-of-the-art local BN structure learning algorithm, which recursively uses standard PC algorithm to find PCs and V-structures.
However, PCD-by-PCD only finds the V-structures connecting the node with its spouses at each iteration, and hence, PCD-by-PCD only finds non-collider V-structures leading to missing some collider V-structures at each iteration.

In the following, under the faithfulness and correct independence tests assumption, we use PCD-by-PCD to find a part of an ALARM~\cite{beinlich1989alarm} BN structure around node 10 to a depth of 2, as illustrated in Fig. 4 (b).
Before giving the example step by step, to make the process easier for readers to understand, as shown in Fig. 5, we first give a detailed description of the three Meek-rules~\cite{meek1995causal} used by PCD-by-PCD in edge orientation step as follows:


\begin{itemize}

\item[\emph{R1}] \emph{No new V-structure.} Orient $Y-Z$ into $Y \rightarrow Z$ whenever there is a directed edge $X \rightarrow Y$ such that $X$ and $Z$ are not adjacent;
\item[\emph{R2}] \emph{Preserve acyclicity.} Orient $X-Z$ into $X \rightarrow Z$ whenever there is a chain $X \rightarrow Y \rightarrow Z$;
\item[\emph{R3}] \emph{Enforce 3-fork V-structure.} Orient $X-Y$ into $X \rightarrow Y$ whenever there are two chains $X- Z \rightarrow Y$ and $X-W \rightarrow Y$ such that $Z$ and $W$ are not adjacent.

\end{itemize}

\begin{figure}[t]
\centering
      \includegraphics[width=3.4in, height=0.75in]{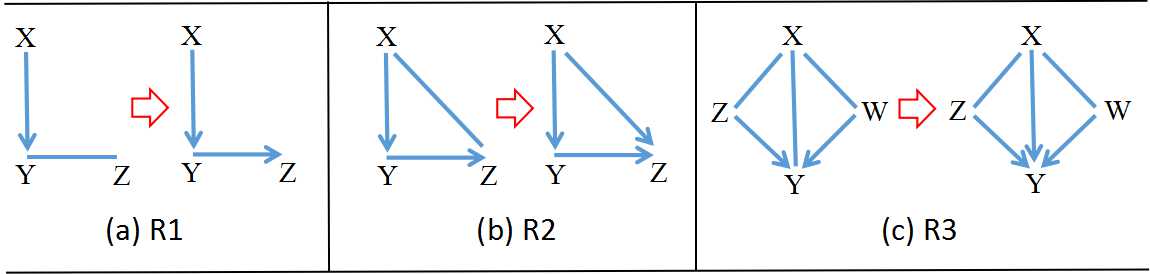}
 \caption{Three Meek-rules for edge orientations.}
 \label{Figure 1}
\end{figure}


\emph{1st iteration:} PCD-by-PCD finds PC of 10. PCD-by-PCD uses symmetry constraint to generate undirected edges, for example, PCD-by-PCD generates undirected edge $A-B$ only if $A$ belongs to the PC of $B$ and B also belongs to the PC of $A$. Since PC of 10 is $\{11,35\}$, but PCs of 11 and 35 are initialized as empty sets and can only be discovered in the next iterations, then 10 does not belong to the PCs of 11 and 35, and there are no undirected edges generated in this iteration.

\emph{2nd iteration:} PCD-by-PCD finds PC of 11. Since PC of 10 is $\{11,35\}$ and PC of 11 is $\{10,12\}$, then 10 belongs to the PC of 11 and 11 also belongs to the PC of 10, and PCD-by-PCD generates undirected edge 10-11. There are no V-structures generated in this iteration, so PCD-by-PCD does not need to orient edges.

\emph{3rd iteration:} PCD-by-PCD finds PC of 35, then generates undirected edge 10-35. Since the non-collider V-structure $11\rightarrow10\leftarrow35$ is discovered, PCD-by-PCD orient the non-collider V-structure, and there are no other undirected edges can be oriented by using Meek-rules.

\emph{4th iteration:} PCD-by-PCD finds PC of 12, then generates undirected edges 12-11 and 12-35. Since PCD-by-PCD only discovers non-collider V-structure at each iteration, it misses the collider V-structure $11\rightarrow12\leftarrow35$. And there are no other undirected edges can be oriented by using Meek-rules.

\emph{5th iteration:} PCD-by-PCD finds PC of 9, and generates undirected edge 9-35. Then there are no new V-structures generated and no other undirected edges can be oriented by using Meek-rules.

\emph{6th-9th iterations:} PCD-by-PCD iteratively finds PCs of 34, 36, 8, and 13, and PCD-by-PCD correctly orients edges in these iterations, so we omit them.

\emph{10th iteration:} PCD-by-PCD finds PC of 15, then generates undirected edge 15-34, and discovers non-collider V-structure $15\rightarrow34\leftarrow13$. Finally, according to the \emph{R1} of Meek-rules, PCD-by-PCD backtracks the edges $34\rightarrow35$, $35\rightarrow12$, $35\rightarrow36$, and $12\rightarrow11$. Thus, the edge $12\rightarrow11$ is falsely oriented.

\textbf{2) Missing non-collider V-structures}: CMB~\cite{gao2015local} is another state-of-the-art local BN structure learning algorithm, which recursively uses standard MB algorithm to find MBs and tracks the conditional independence changes to find V-structures.
However, CMB only finds the V-structures included in the PC of the target at each iteration. Thus, CMB only finds collider V-structures and then misses some non-collider V-structures at each iteration.

In the following, under the faithfulness and correct independence tests assumption, we use CMB to find a part of an ALARM BN structure around node 26 to a depth of 2, as illustrated in Fig. 4 (c). Moreover, CMB tracks the conditional independence changes in edge orientation step, which is similar to the three Meek-rules~\cite{gao2015local}.


\emph{1st iteration:} CMB finds MB of 26 and generates undirected edges using PC of 26. Then CMB discovers the collider V-structures $25\rightarrow26\leftarrow30$, $30\rightarrow26\leftarrow17$, and $25\rightarrow26\leftarrow17$, and there are no other undirected edges can be oriented by tracking the conditional independence changes.

\emph{2nd iteration:} CMB finds MB of 17, then generates undirected edge 17-31, and there are no other undirected edges can be oriented by tracking the conditional independence changes.

\emph{3rd iteration:} CMB finds MB of 25 and generates undirected edges using PC of 25. Since CMB only finds collider V-structures at each iteration, it misses the non-collider V-structure $25\rightarrow31\leftarrow17$. Then there are no other undirected edges can be oriented by tracking the conditional independence changes.

\emph{4th iteration:} CMB finds MB of 30 and generates undirected edges using PC of 30. Since CMB discovers collider V-structure $27\rightarrow30\leftarrow29$, CMB orients the collider V-structure. Then according to the conditional independence changes, CMB executes the same way as the \emph{R1} of the Meek-rules to backtrack the edges $30\rightarrow31$, $31\rightarrow25$, $31\rightarrow17$, $25\rightarrow18$, $25\rightarrow24$, and $25\rightarrow32$. Thus, the edges $31\rightarrow25$ and $31\rightarrow17$ are falsely oriented.

\textbf{Summary}: Local BN structure learning algorithms miss V-structures in Expand-Backtracking, and thus they encounter the false edge orientation problem when learning any part of a BN structure.
If we do not tackle the missing V-structures in Expand-Backtracking, many edges may be falsely oriented during the edge orientation step, leading to low accuracy of any part of BN structure learning.

Clearly, to tackle the missing V-structures in Expand-Backtracking when learning any part of a BN structure, we need to correctly identify both of non-collider V-structures and collider V-structures in the current part of a BN skeleton at each iteration.

\section{The proposed APSL and APSL-FS algorithms}

This section presents the proposed any part of BN structure learning algorithms, APSL in Section V-A and APSL-FS in Section V-B.

\subsection{APSL: Any Part of BN Structure Learning}

With the analysis of missing V-structures in Expand-Backtracking in Section IV, we present the proposed APSL (\underline{A}ny \underline{P}art of BN \underline{S}tructure \underline{L}earning) algorithm, as described in Algorithm 1. APSL recursively finds both of non-collider V-structures (Step 1: Lines 9-26) and collider V-structures (Step 2: Lines 28-36) in MBs, until all edges in the part of a BN structure around the target node are oriented (Step 3: Lines 38-58).

\begin{figure}[!htbp]
\centering
 \includegraphics[width=3.3in, height=0.6in]{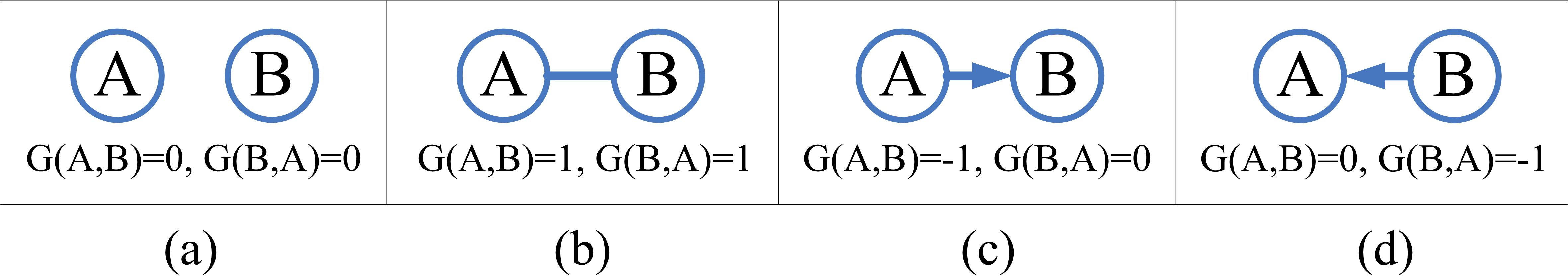}
 \caption{Four types of relationships between two variables.}
 \label{Figure 1}
\end{figure}

APSL defines an adjacency matrix $G$ of a DAG, to detect the relationship among all the variables.
In $G$, the four types of the relationship between any two variables $A$ and $B$ are shown in Fig. 6, as follows:

\begin{enumerate}

\item[(a)] $A$ and $B$ are not adjacent $\Rightarrow$ $G(A,B)=0$ and $G(B,A)=0$.
\item[(b)] $A$ and $B$ are adjacent but cannot determine their edge direction $\Rightarrow$ $G(A,B)=1$ and $G(B,A)=1$.
\item[(c)] $A$ and $B$ are adjacent and $A\rightarrow B$ $\Rightarrow$ $G(A,B)=-1$ and $G(B,A)=0$.
\item[(d)] $A$ and $B$ are adjacent and $A\leftarrow B$ $\Rightarrow$ $G(A,B)=0$ and $G(B,A)=-1$.

\end{enumerate}

\begin{algorithm}[!htbp]
\small
    \caption{APSL}

    \label{alg:r2p}
    \KwIn{$\mathcal{D}$: Data, $T$: Target, $K$: a given depth\;}
    \KwOut{$G$: a part of a BN structure around $T$\;}
    $\textbf{V}=\emptyset$\;
    $Q=\{T\}$\;

    $G = zeros(|\textbf{U}|,|\textbf{U}|)$\;

    $layer\_num=1$\;

    $\textbf{layer\_nodes}(layer\_num)=T$\;

    $i=1$\;

    \Repeat{\rm $|\textbf{V}|=|\textbf{U}|$, or $Q=\emptyset$}
    {

           /*$Step\ 1:\ Find\ non$-$collider V$-$structures$*/

           $A=Q.pop$\;
           \If{\rm $A\in \textbf{V}$}
           {
               continue;
           }

           $[\textbf{PC}_{A},\textbf{SP}_{A}]=$ GetMB$(\mathcal{D}, A)$

           $\textbf{V}=\textbf{V}\cup \{A\}$\;

           $Q.push(\textbf{PC}_{A})$\;

          \For{\rm each $B\in\textbf{PC}_{A}$}
          {

             \If{$G(A,B)=0 \& G(B,A)=0$}
             {
                $G(A,B)=1, G(B,A)=1$\;
             }
           }

           \For{\rm each $B\in\textbf{PC}_{A}$}
           {
               \For{\rm each $C\in\textbf{SP}_{A}(B)$}
                {
                   $G(A,B)=-1, G(B,A)=0$\;
                   $G(C,B)=-1, G(B,C)=0$\;
                }
           }

           /*$Step\ 2:\ Find\ collider\ V$-$structures $*/

           \For{\rm every $X,Y\in\textbf{PC}_{A}$}
           {
                \If{\rm $X\!\perp\!\!\!\perp Y|\textbf{Z}$ for some $\textbf{Z}\subseteq\textbf{PC}_{X}$}
                {
                        $\textbf{Sep}_{X}[Y]=\textbf{Z}$\;
                        \If{\rm $X\not\!\perp\!\!\!\perp Y|\textbf{Sep}_{X}[Y]\cup \{A\}$}
                        {
                            $G(X,A)=-1, G(A,X)=0$\;
                            $G(Y,A)=-1, G(A,Y)=0$\;
                        }
                 }

           }

           /*$Step\ 3:\ Orient\ edges\ $*/

           update $G$ by using Meek rules \;

           $i=i-1$\;

           \If{$i=0$}
           {
                $layer\_num=layer\_num+1$\;

                \For{\rm each $X\in\textbf{layer\_nodes}(layer\_num-1)$}
                {

                    $\textbf{layer\_nodes}(layer\_num)=\textbf{layer\_nodes}(layer\_num)\cup \textbf{PC}_{X}$\;
                }

                $i=|\textbf{layer\_nodes}(layer\_num)\setminus\textbf{V}|$\;
           }

           \If{$layer\_num > K$}
           {
                $break\_flag$=1;

                \For{\rm each $X\in\textbf{layer\_nodes}(K)$}
                {
                    \If{\rm can find $G(\textbf{PC}_{X},X)=1$}
                    {
                        $break\_flag$=0\;
                        \textbf{break}\;
                    }
                }
                \If{\rm $break\_flag$}
                {
                    \textbf{break}\;
                }
           }
    }

    Return $G$\;

\end{algorithm}

APSL first initializes the queried variable set $\textbf{V}$ to an empty set and initializes the queue $Q$, pre-storing the target variable $T$. Then, the next three steps will be repeated until all edges in the part of a BN structure around $T$ to a depth of $K$ are oriented, or the size of $\textbf{V}$ equals to that of the entire variable set $\textbf{U}$, or $Q$ is empty.

\textbf{Step 1: Find non-collider V-structures} (Lines 9-26). APSL first pops the first variable $A$ from the queue $Q$, and then uses MB discovery algorithms to find the MB (i.e., PC and spouse) of $A$. APSL will first find the PC and spouse of $T$ since $T$ is pre-stored in $Q$. Then, APSL pushes the PC of $A$ into $Q$ to recursively find the MB of each node in the PC of $A$ in the next iterations, and stores $A$ in $\textbf{V}$ to prevent repeated learning.
Finally, APSL generates undirected edges by using the PC of $A$ (Lines 16-20), and orients the non-collider V-structures by using the spouses of $A$ (Lines 21-26).

At Line 13, the MB discovery algorithm, we use is a constraint-based MB method, such as MMMB~\cite{tsamardinos2003time} or HITON-MB~\cite{aliferis2003hiton}, because this type of MB methods do not require a lot of memory. Moreover, these MB methods can save the discovered PCs to avoid repeatedly learning PC sets during any part of BN structure learning, since they find spouses from the PC of each variable in the target's PC.
Line 17 aims to prevent the already oriented edges from being re-initialized as undirected edges. $layer\_num$ represents the number of layers, starting from 1. Thus, the number of layers is one more than the corresponding number of depths, for example, when the number of depths is 2, the corresponding number of layers is 3. \textbf{layer\_nodes} stores the nodes of each layer.

\textbf{Step 2: Find collider V-structures} (Lines 28-36). APSL finds collider V-structures in the PC of $A$. If two variables $X$ and $Y$ in the PC of $A$ are conditionally independent, that is, they are not adjacent owing to Theorem 1. But these two variables are conditionally dependent given the union of the collider $A$ and their separating set, then the triple of nodes $X$, $Y$, and $A$ can form collider V-structure of $A$ owing to Theorem 2, $X\rightarrow A\leftarrow Y$.

\textbf{Step 3: Orient edges} (Lines 38-58). Based on the oriented non-collider V-structures and collider V-structures, APSL uses Meek-rules to orient the remaining undirected edges (Line 38).
The purpose of Lines 40-46 is to control the number of layers of recursion. Specifically, $i$ reduced by 1 at each iteration, and $i=0$ means that all the nodes in this layer have been traversed, then ASPL begins to traverse the nodes at the next layer in the next iterations.
From Lines 47-58, APSL determines whether all edges in the part of a BN structure around $T$ are oriented. When the edges between the layer of $K$ and $K$+1 of a part of a BN structure around $T$ are all oriented, APSL terminates and outputs the part of a BN structure around $T$. Some edges with a number of layers less than $K$ are not oriented because these edges can never be oriented due to the existence of Markov equivalence structures~\cite{xie2006decomposition}.


\textbf{Theorem 4 \emph{Correctness of APSL}} Under the faithfulness and correct independence tests assumption, APSL finds a correct part of a BN structure.

\emph{\textbf{Proof}}
Under the faithfulness and correct independence tests assumption, we will prove the correctness of APSL in three steps.

\emph{1) Step 1 finds all and only the non-collider V-structures.} A standard MB discovery algorithm finds all and only the PC and spouses of a target node. APSL uses the MB method to find PC and spouses of the nodes that need to be found. Then, using the found PCs, APSL constructs a part of a BN skeleton with no missing edges and no extra edges. Using the found spouses, APSL finds all and only the non-collider V-structures.

\emph{2) Step 2 finds all and only the collider V-structures.}
APSL finds collider V-structures in PCs. First, APSL uses Theorem 1 to confirm that there is no edge between two nodes $X$ and $Y$ in the PC of $A$ (the target node at each iteration). Then, owing to Theorem 2, if the collider $A$ makes $X$ and $Y$ conditionally dependent, $X\not\!\perp\!\!\!\perp Y|\textbf{Sep}_{X}[Y]\cup \{A\}$, then $X$ and $Y$ are each other's spouses with the common child $A$, and forms a collider V-structure $X\rightarrow A\leftarrow Y$. Since APSL considers any two nodes in the PCs and their common child, APSL finds all and only the collider V-structures.

\emph{3) Step 3 finds a correct part of a BN structure.}
Based on the part of a BN skeleton with all non-collider V-structures and collider V-structures, APSL uses Meek-rules to recover the part of a skeleton to a correct part of a structure, some edges cannot be oriented due to the existence of Markov equivalence structures.
Finally, APSL terminates when the part of a structure expands to a given depth, and thus APSL finds a correct part of a BN structure. \hfill$\blacksquare$

\begin{figure}[t]
\centering
       \begin{tabular}{c|c}

      \subfigure{\includegraphics[width=0.93in, height=2.4in]{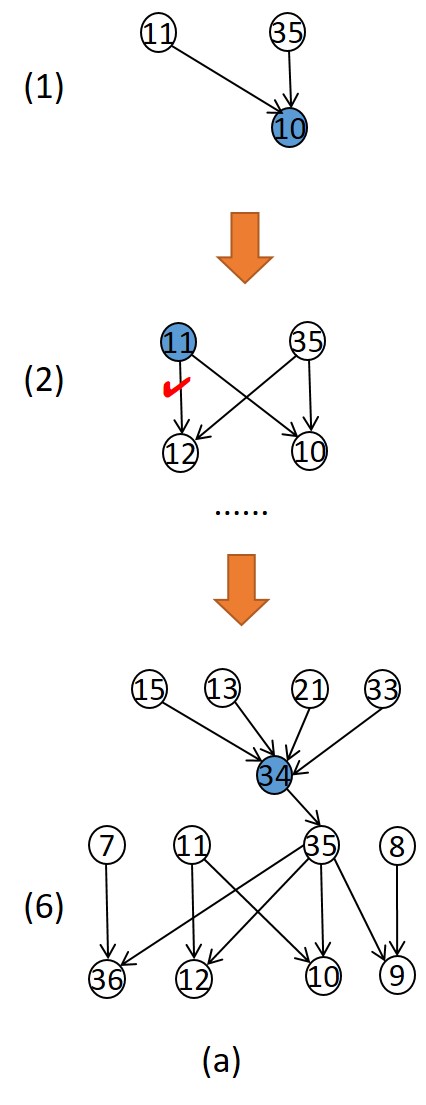}}&
      \subfigure{\includegraphics[width=2.2in, height=2.4in]{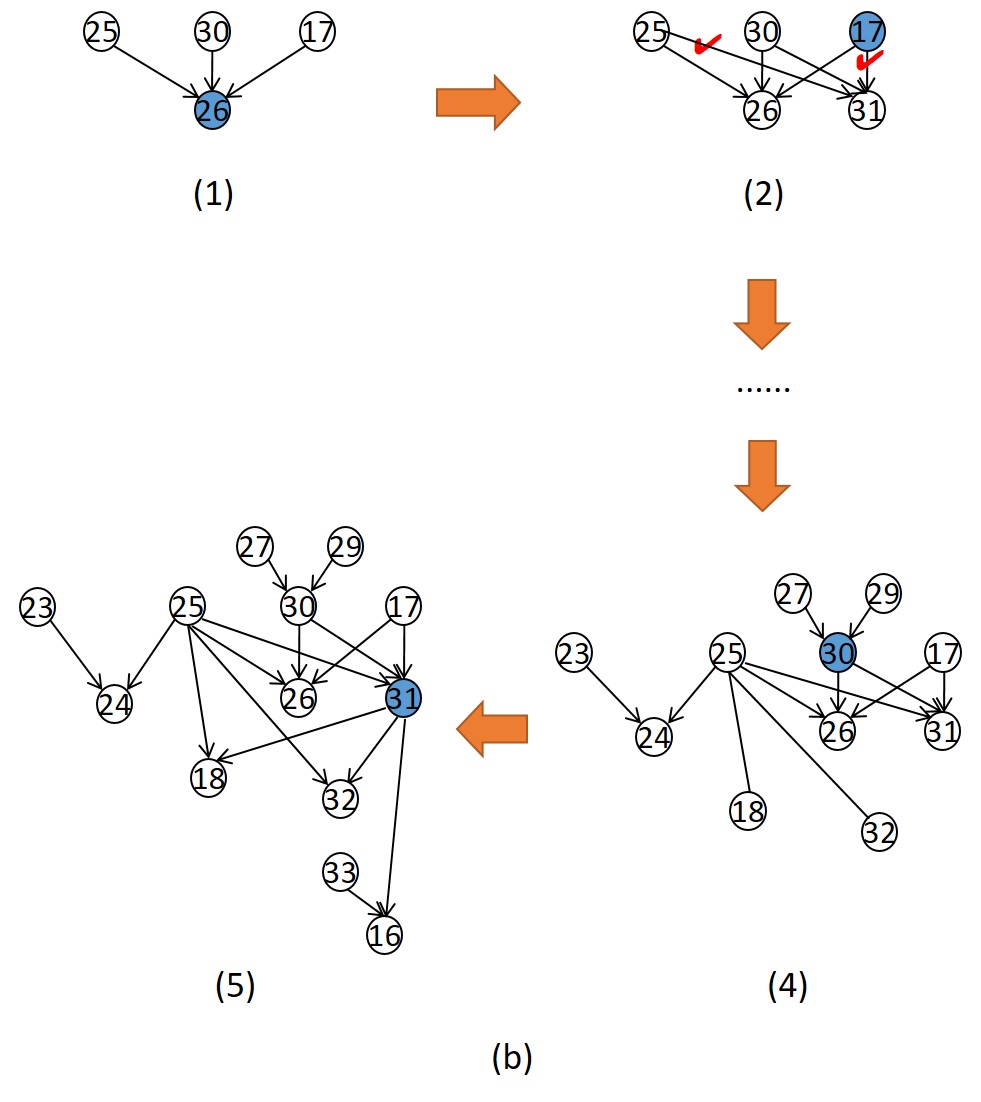}}\\

       \end{tabular}
 \caption{(a) An example of using APSL to find a part of an Alarm Bayesian network structure around node 10 to a depth of 2; (b) an example of using APSL to find a part of an Alarm Bayesian network structure around node 26 to a depth of 2.
 \protect\\The red '\checkmark' symbol denotes the edges that local BN structure learning algorithm falsely orients but APSL correctly orients, the blue node is the target node during each iteration, the number in parentheses represents the level of iterations, and '$\cdots$' means omitted iterations.}
 \label{Figure 1}
\end{figure}

\textbf{Tracing APSL} To further validate that our algorithm can tackle missing V-structures in Expand-Backtracking, we use the same examples in Fig. 4 to trace the execution of APSL.

\emph{Case 1:} As shown in Fig. 7 (a), APSL finds the collider V-structure of 10 at the 1st iteration, $11\rightarrow10\leftarrow35$. Then, at the 2nd iteration, APSL finds the non-collider V-structure of 11, $11\rightarrow12\leftarrow35$, which is missed by PCD-by-PCD.

\emph{Case 2:} As shown in Fig. 7 (b), At the 1st iteration, APSL finds the collider V-structures of 26. And at the 2nd iteration, APSL finds the non-collider V-structure of 17, $25\rightarrow26\leftarrow17$, which is missed by CMB.

\subsection{APSL-FS: APSL using Feature Selection}

In this section, we will propose an efficient version of APSL by using feature selection.

APSL uses a standard MB discovery algorithm, MMMB or HITON-MB, for MB discovery. However, the standard PC discovery algorithms, MMPC~\cite{tsamardinos2003time} and HITON-PC~\cite{aliferis2003hiton} (used by MMMB and HITON-MB, respectively), need to perform an exhaustive subset search within the currently selected variables as conditioning sets for PC discovery, and thus they are computationally expensive or even prohibitive when the size of the PC set of the target becomes large.

Feature selection is a common dimensionality reduction technique and plays an essential role in data analytics~\cite{guyon2003introduction,yu2016scalable,yu2019MCFS}.
Existing feature selection methods can be broadly categorized into embedded methods, wrapper methods, and filter methods~\cite{wu2013online}. Since filter feature selection methods are fast and independent of any classifiers, they have attracted more attentions.

It has been proven in our previous work~\cite{yu2018unified} that some filter feature selection methods based on mutual information prefer the PC of the target variable. Furthermore, these methods use pairwise comparisons~\cite{yu2004efficient} (i.e., unconditional independence tests) to remove false positives with less correlations, they can find the potential PC of the target variable without searching for conditioning set, and thus improving the efficiency of PC discovery.

Thus, to address the problem exists in APSL for PC discovery, we use a filter feature selection method based on mutual information instead of the standard PC discovery algorithm. However, the feature selection method we use cannot find spouses for edge orientations. Because the feature selection method uses pairwise comparisons rather than conditional independence tests~\cite{yu2004efficient}, it cannot find the separating sets which is the key to finding spouses~\cite{aliferis2010local1}.

Standard PC discovery algorithms find separating sets to make a target variable and the other variables conditionally independent, only the variables in the PC of the target are always conditionally dependent on the target~\cite{aliferis2010local1}. Thus, standard PC discovery algorithms find PC and separating sets simultaneously. However, these algorithms are computationally expensive in finding separating sets since they need to find the separating sets of all variables independent of the target. Instead, it is only necessary to find the separating sets of the variables in the PC of each variable in the target's PC set, as spouses of the target variable exist only there.

Thus in this subsection, based on using feature selection for PC discovery, we propose an efficient Markov blanket discovery algorithm for spouses discovery, called MB-FS (\underline{M}arkov \underline{B}lanket discovery by \underline{F}eature \underline{S}election). Moreover, we use MB-FS instead of the standard MB discovery algorithm for MB discovery in APSL to improve the efficiency, and we call this new any part of BN structure learning algorithm APSL-FS (APSL using \underline{F}eature \underline{S}election), an efficient version of APSL using feature selection. In the following, we will go into details about using feature selection for PC discovery and MB discovery, respectively.

\begin{algorithm}[t]
\small
    \caption{FCBF}
    \label{alg:r2p}
    \KwIn{$\mathcal{D}$: Data, $T$: Target, $\delta$: Threshold\;}
    \KwOut{$\textbf{PC}_{T}$: PC of $T$ \;}


    $\textbf{S}=\emptyset$\;

    \For{\rm each $X\in \textbf{U}\setminus\{T\}$}
    {

      \If{$SU(X;T)>\delta$}
      {
        $\textbf{S}=\textbf{S}\cup X$\;
      }
    }


    Order \textbf{S} in descending $SU(X;T)$ value;

    $i=1$\;

    \While{$i<=|\textbf{S}|$}
    {
        $j=i+1$\;

        \While{$j<=|\textbf{S}|$}
        {

            \eIf{$SU(\textbf{S}(i);\textbf{S}(j))>SU(\textbf{S}(j);T)$}
            {
                $\textbf{S}(j)=\emptyset$\;
            }
            {
                $j=j+1$\;
            }

        }
         $i=i+1$\;
    }

    $\textbf{PC}_{T}=\textbf{S}$\;

    Return $\textbf{PC}_{T}$\;
\end{algorithm}

\begin{algorithm}[t]
\small
    \caption{MB-FS}
    \label{alg:r2p}
    \KwIn{$\mathcal{D}$: Data, $T$: Target, $\delta$: Threshold\;}
    \KwOut{[$\textbf{PC}_{T},\textbf{SP}_{T}$]: MB of $T$ \;}

    $\textbf{PC}_{T} =  \rm FCBF(\mathcal{D}, \emph{T}, \delta)$\;

    \For{\rm each $X\in\textbf{PC}_{T}$}
    {
      $\textbf{PC}_{X} = \rm FCBF(\mathcal{D}, \emph{X}, \delta)$\;
      \For{\rm each $Y\in\textbf{PC}_{X}$}
      {

        \If{\rm $T\!\perp\!\!\!\perp Y|\textbf{Z}$ for some $\textbf{Z}\subseteq\textbf{PC}_{T}$}
        {
            $\textbf{Sep}_{T}[Y]=\textbf{Z}$\;
            \If{\rm $T\not\!\perp\!\!\!\perp Y|\textbf{Sep}_{T}[Y]\cup \{X\}$}
            {
                $\textbf{SP}_{T}(X)=\textbf{SP}_{T}(X)\cup \{Y\}$\;
            }
        }
      }
    }
    Return [$\textbf{PC}_{T},\textbf{SP}_{T}$]\;
\end{algorithm}

\emph{(1) PC discovery:} We choose a well-established feature selection method, Fast Correlation-Based Filter (FCBF)~\cite{yu2004efficient}, for PC discovery because the size of the PC of each variable in a BN is not fixed. FCBF specifies a threshold $\delta$ to control the number of potential PC of the target variable, instead of specifying the number of the PC in advance.

As illustrated in Algorithm 2, FCBF first finds a potential PC of the target variable from the entire variable set whose correlations with the target are higher than the threshold (Lines 1-6). Then, FCBF uses pairwise comparisons to remove false positives in the potential PC to get the true PC (Lines 7-20).


\emph{(2) MB discovery:} As illustrated in Algorithm 3, MB-FS first uses FCBF to find the PC of the target variable $T$, and uses FCBF to find the PC of each variable in the $T$'s PC as the candidate spouses of $T$. Then, MB-FS finds the separating set from the subsets of the PC of $T$, to make $T$ and the variable $Y$ in the candidate spouses are conditionally independent. Finally, if $T$ and $Y$ are conditionally dependent given the union of the separating set and their common child $X$, $Y$ is a spouse of $T$ owing to Theorem 2.

\section{Experiments}


In this section, we will systematically evaluate our presented algorithms. In Section VI-A, we describe the data sets, comparison methods, and evaluation metrics in the experiments. Then in Section VI-B and VI-C, we evaluate our algorithms with local BN structure learning algorithms and global BN structure learning algorithms, respectively.

\begin{table}[!htbp]
\caption{Summary of benchmark BNs }{
\footnotesize
\begin{center}
\begin{tabular}{ccccc}
\toprule

            & Num. & Num.  & Max In/out-     & Min/Max     \\
Network     & Vars & Edges & Degree          & $|$PCset$|$ \\

\midrule


Child       & 20   & 25    & 2/7          & 1/8     \\
Insurance   & 27   & 52    & 3/7          & 1/9     \\
Alarm       & 37   & 46    & 4/5          & 1/6     \\

Child10     & 200  & 257   & 2/7          & 1/8     \\
Insurance10 & 270  & 556   & 5/8          & 1/11    \\
Alarm10     & 370  & 570   & 4/7          & 1/9     \\

%
%
%

\bottomrule
\end{tabular}
\end{center}}
\end{table}

\subsection{Experiment setting}


To evaluate the APSL and APSL-FS algorithms, we use two groups of data generated from the six benchmark BNs as shown in Table II\footnote{The public data sets are available at $http://pages.mtu.edu/\sim lebrown/supplements/mmhc\_paper/mmhc\_index.html$.}. One group includes 10 data sets each with 500 data instances, and the other group also contains 10 data sets each with 1,000 data instances.

We compare the APSL and APSL-FS algorithms with 7 other algorithms, including 2 local BN structure learning algorithms, PCD-by-PCD~\cite{yin2008partial} and CMB~\cite{gao2015local}, and 5 global BN structure learning algorithms, GS~\cite{margaritis2000bayesian}, MMHC~\cite{tsamardinos2006max}, SLL+C~\cite{niinimki2012local}, SLL+G~\cite{niinimki2012local}, and GGSL~\cite{gao2017local}.

The implementation details and parameter settings of all the algorithms are as follows:

\begin{enumerate}

\item PCD-by-PCD, CMB, GS, MMHC\footnote{The codes of MMHC in MATLAB are available at $http://mensxmachina.org/en/software/probabilistic-graphical-model-toolbox/$.}, APSL, and APSL-FS are implemented in MATLAB, SLL+C/G\footnote{The codes of SLL+C and SLL+G in C++ are available at $https://www.cs.helsinki.fi/u/tzniinim/uai2012/$.} and GGSL are implemented in C++.

\item The conditional independence tests are $G^{2}$ tests with the statistical significance level of 0.01, the constrained MB algorithm used by APSL is HITON-MB~\cite{aliferis2003hiton}, and the threshold of the feature selection method FCBF~\cite{yu2004efficient} used by APSL-FS is 0.05.

\item In all Tables in Section VI, the experimental results are shown in the format of $A\pm B$, where $A$ represents the average results, and $B$ is the standard deviation. The best results are highlighted in boldface. 

\item All experiments are conducted on a computer with an Intel Core i7-8700 3.20 GHz with 8GB RAM.

\end{enumerate}

%
%
%

\begin{table*}[t]
\centering
\footnotesize
\caption{Ar\_Distance, Ar\_Precision, Ar\_Recall, and Runtime (in seconds) on learning a part of BN structures to a depth of 1 using different data sizes ($A\pm B$: $A$ represents the average results while $B$ is the standard deviation. The best results are highlighted in boldface.)}
\begin{tabular}{cc||cccc|cccc}

\hline
       &           & \multicolumn{2}{l}{Size=500}   &           &                    & \multicolumn{2}{l}{Size=1,000}  &           &                     \\
Network                      & Algorithm  & Ar\_Distance    & Ar\_Precision & Ar\_Recall & Runtime               & Ar\_Distance    & Ar\_Precision & Ar\_Recall & Runtime               \\\hline
\multirow{4}{*}{Child}       & PCD-by-PCD & 0.57$\pm$0.55          & 0.66$\pm$0.42        & 0.56$\pm$0.39     & 0.17$\pm$0.08          & 0.46$\pm$0.47          & 0.69$\pm$0.35        & 0.67$\pm$0.34     & 0.25$\pm$0.14          \\
                             & CMB        & 0.54$\pm$0.52          & 0.61$\pm$0.38        & 0.65$\pm$0.39     & 0.63$\pm$0.24          & 0.39$\pm$0.50          & 0.73$\pm$0.36        & 0.73$\pm$0.36     & 0.67$\pm$0.38          \\
                             & APSL       & 0.63$\pm$0.48          & 0.55$\pm$0.34        & 0.57$\pm$0.37     & 0.12$\pm$0.06          & 0.45$\pm$0.50          & 0.70$\pm$0.36        & 0.68$\pm$0.36     & 0.18$\pm$0.06          \\
                             & ASPL-FS    & \textbf{0.37$\pm$0.46} & 0.80$\pm$0.32        & 0.70$\pm$0.35     & \textbf{0.03$\pm$0.02} & \textbf{0.37$\pm$0.46} & 0.83$\pm$0.32        & 0.70$\pm$0.35     & \textbf{0.03$\pm$0.03} \\\hline
\multirow{4}{*}{Insurance}   & PCD-by-PCD & 0.90$\pm$0.38          & 0.45$\pm$0.35        & 0.33$\pm$0.28     & 0.16$\pm$0.15          & 0.70$\pm$0.46          & 0.64$\pm$0.38        & 0.43$\pm$0.32     & 0.23$\pm$0.20          \\
                             & CMB        & 0.81$\pm$0.44          & 0.49$\pm$0.36        & 0.40$\pm$0.33     & 0.36$\pm$0.22          & 0.77$\pm$0.43          & 0.52$\pm$0.34        & 0.42$\pm$0.31     & 0.56$\pm$0.37          \\
                             & APSL       & 0.83$\pm$0.38          & 0.48$\pm$0.34        & 0.38$\pm$0.25     & 0.10$\pm$0.09          & \textbf{0.51$\pm$0.36} & 0.76$\pm$0.30        & 0.59$\pm$0.27     & 0.14$\pm$0.10          \\
                             & ASPL-FS    & \textbf{0.66$\pm$0.50} & 0.67$\pm$0.43        & 0.48$\pm$0.33     & \textbf{0.05$\pm$0.06} & 0.85$\pm$0.48          & 0.49$\pm$0.42        & 0.36$\pm$0.32     & \textbf{0.07$\pm$0.06} \\\hline
\multirow{4}{*}{Alarm}       & PCD-by-PCD & 0.61$\pm$0.57          & 0.60$\pm$0.42        & 0.56$\pm$0.41     & 0.21$\pm$0.22          & 0.50$\pm$0.55          & 0.69$\pm$0.40        & 0.63$\pm$0.40     & 0.29$\pm$0.27          \\
                             & CMB        & 0.65$\pm$0.59          & 0.58$\pm$0.44        & 0.53$\pm$0.42     & 0.32$\pm$0.33          & 0.58$\pm$0.58          & 0.62$\pm$0.42        & 0.58$\pm$0.41     & 0.24$\pm$0.19          \\
                             & APSL       & \textbf{0.53$\pm$0.57} & 0.66$\pm$0.42        & 0.63$\pm$0.40     & 0.17$\pm$0.16          & \textbf{0.41$\pm$0.49} & 0.75$\pm$0.36        & 0.70$\pm$0.36     & 0.23$\pm$0.20          \\
                             & ASPL-FS    & 0.61$\pm$0.55          & 0.61$\pm$0.41        & 0.55$\pm$0.40     & \textbf{0.07$\pm$0.09} & 0.51$\pm$0.55          & 0.69$\pm$0.41        & 0.62$\pm$0.39     & \textbf{0.07$\pm$0.09} \\\hline
\multirow{4}{*}{Child10}     & PCD-by-PCD & 0.91$\pm$0.49          & 0.38$\pm$0.38        & 0.35$\pm$0.35     & 2.03$\pm$3.29          & 0.73$\pm$0.55          & 0.51$\pm$0.41        & 0.48$\pm$0.40     & 2.46$\pm$3.73          \\
                             & CMB        & 0.60$\pm$0.47          & 0.60$\pm$0.36        & 0.60$\pm$0.36     & 2.41$\pm$2.61          & 0.55$\pm$0.49          & 0.62$\pm$0.36        & 0.63$\pm$0.36     & 1.98$\pm$2.06          \\
                             & APSL       & 0.68$\pm$0.49          & 0.52$\pm$0.37        & 0.56$\pm$0.38     & 0.83$\pm$1.62          & 0.48$\pm$0.45          & 0.66$\pm$0.33        & 0.69$\pm$0.34     & 1.05$\pm$2.16          \\
                             & ASPL-FS    & \textbf{0.53$\pm$0.53} & 0.70$\pm$0.40        & 0.59$\pm$0.38     & \textbf{0.08$\pm$0.06} & \textbf{0.47$\pm$0.52} & 0.75$\pm$0.39        & 0.63$\pm$0.37     & \textbf{0.15$\pm$0.11} \\\hline
\multirow{4}{*}{Insurance10} & PCD-by-PCD & 0.85$\pm$0.41          & 0.45$\pm$0.33        & 0.38$\pm$0.31     & 1.33$\pm$4.15          & 0.72$\pm$0.43          & 0.58$\pm$0.36        & 0.44$\pm$0.30     & 1.22$\pm$3.93          \\
                             & CMB        & 0.82$\pm$0.39          & 0.46$\pm$0.31        & 0.42$\pm$0.31     & 1.49$\pm$1.30          & 0.68$\pm$0.41          & 0.57$\pm$0.33        & 0.50$\pm$0.31     & 2.06$\pm$1.84          \\
                             & APSL       & \textbf{0.71$\pm$0.38} & 0.54$\pm$0.31        & 0.50$\pm$0.30     & \textbf{0.50$\pm$1.23} & \textbf{0.54$\pm$0.40} & 0.67$\pm$0.31        & 0.61$\pm$0.31     & 0.84$\pm$2.56          \\
                             & ASPL-FS    & \textbf{0.71$\pm$0.40} & 0.66$\pm$0.37        & 0.42$\pm$0.29     & 1.09$\pm$1.86          & 0.60$\pm$0.40          & 0.79$\pm$0.35        & 0.48$\pm$0.30     & \textbf{0.34$\pm$0.32} \\\hline
\multirow{4}{*}{Alarm10}     & PCD-by-PCD & 0.83$\pm$0.51          & 0.50$\pm$0.43        & 0.38$\pm$0.36     & 5.55$\pm$9.42          & 0.72$\pm$0.53          & 0.59$\pm$0.42        & 0.45$\pm$0.37     & 4.91$\pm$9.64          \\
                             & CMB        & 0.81$\pm$0.53          & 0.47$\pm$0.42        & 0.42$\pm$0.39     & 1.42$\pm$1.48          & 0.70$\pm$0.55          & 0.57$\pm$0.42        & 0.49$\pm$0.40     & \textbf{0.92$\pm$0.87} \\
                             & APSL       & 0.70$\pm$0.48          & 0.56$\pm$0.39        & 0.49$\pm$0.36     & 3.16$\pm$6.21          & \textbf{0.55$\pm$0.48} & 0.69$\pm$0.37        & 0.59$\pm$0.36     & 5.19$\pm$8.97          \\
                             & ASPL-FS    & \textbf{0.68$\pm$0.48} & 0.61$\pm$0.39        & 0.49$\pm$0.35     & \textbf{0.82$\pm$1.86} & 0.65$\pm$0.51          & 0.65$\pm$0.41        & 0.49$\pm$0.36     & 1.48$\pm$2.64          \\\hline
\end{tabular}

\label{tab:plain}
\end{table*}

\begin{figure*}[t]
\centering
       \begin{tabular}{ccccc}

      \subfigure{\includegraphics[width=2.8in, height=1.7in]{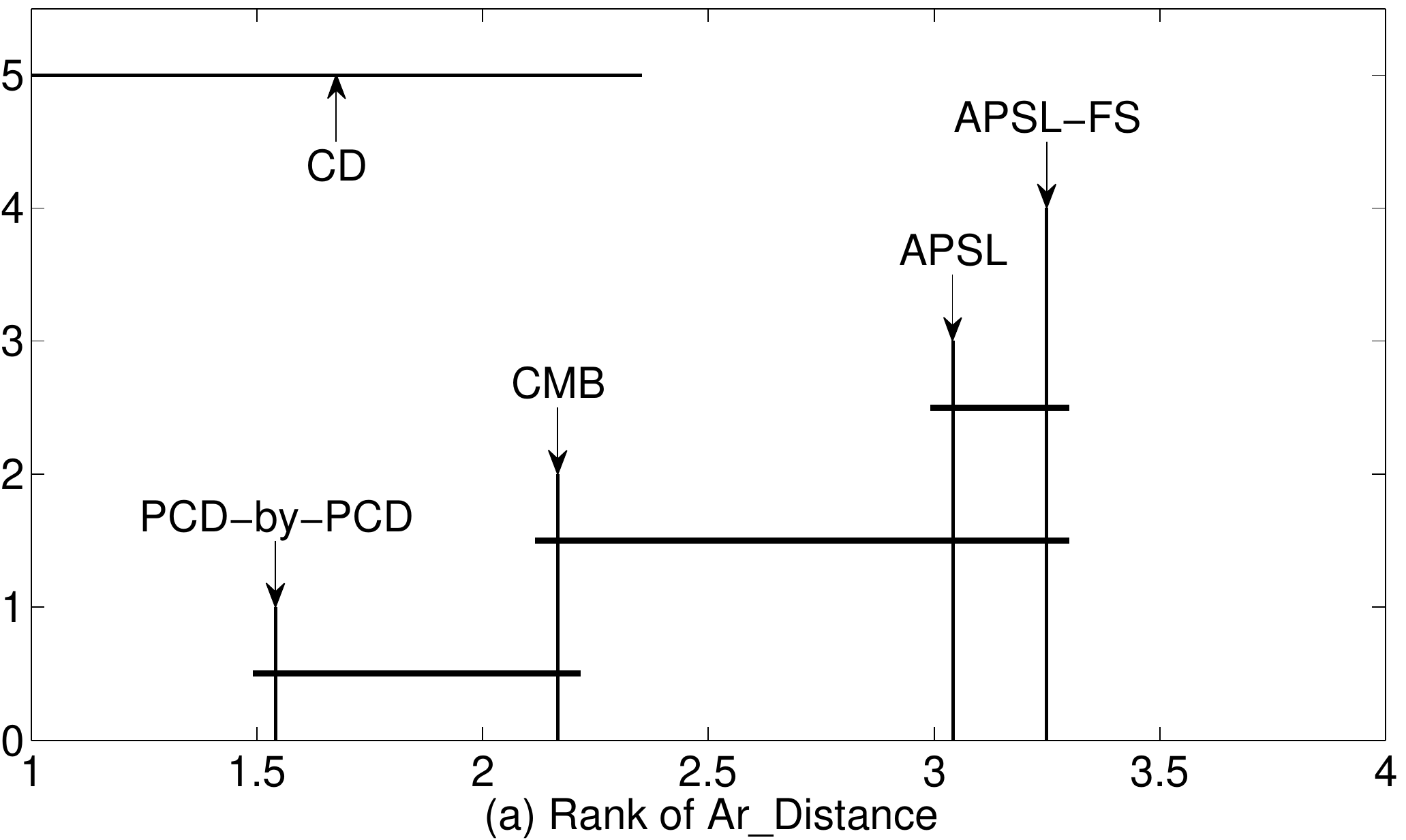}}&
      \subfigure{\includegraphics[width=2.8in, height=1.7in]{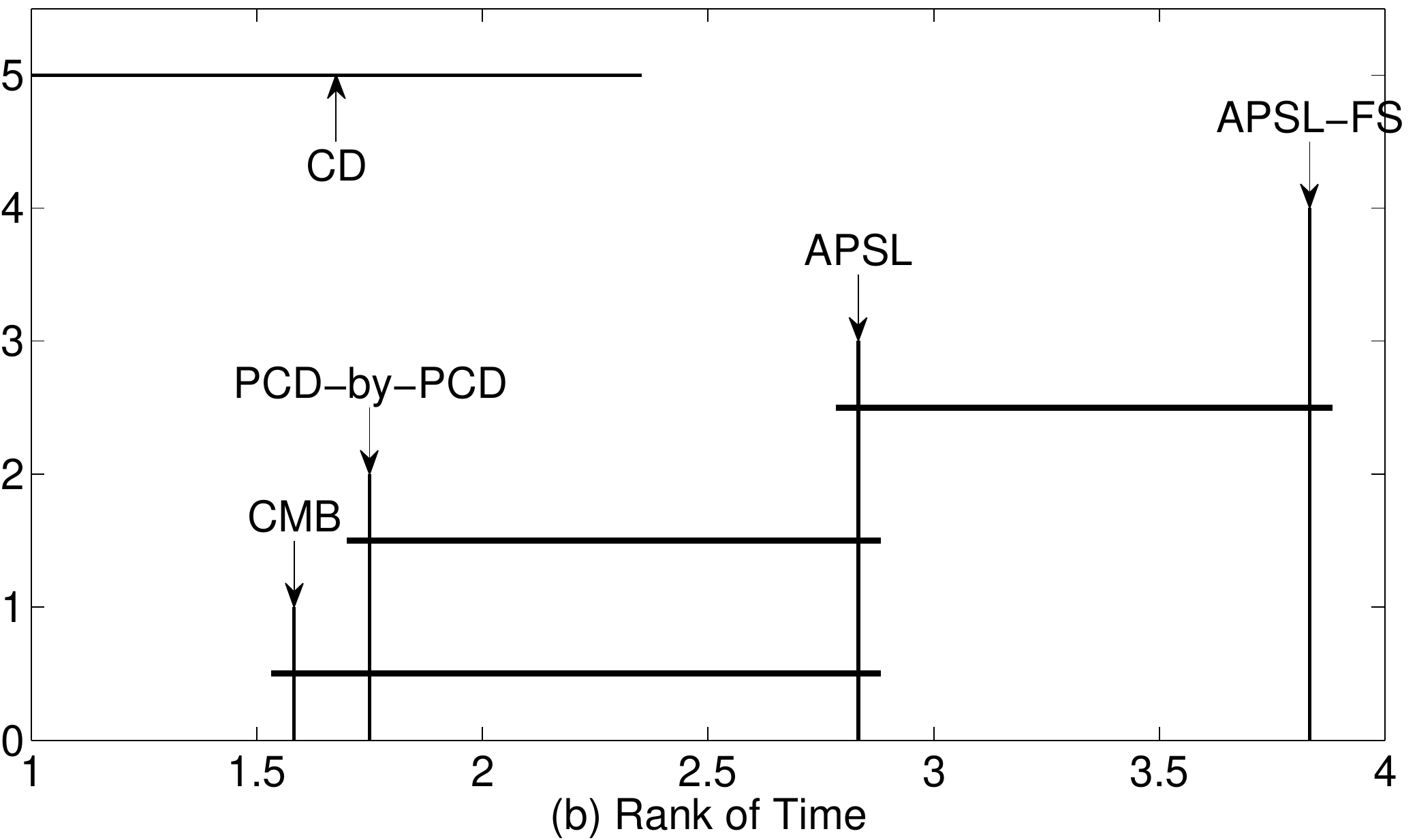}}&

       \end{tabular}
 \caption{Crucial difference diagram of the Nemenyi test of Ar\_Distance and Runtime on learning a part of BN structures (Depth=1).}
 \label{Figure 9}
\end{figure*}


Using the BN data sets, we evaluate the algorithms using the following metrics:

\begin{itemize}

\item Accuracy. We evaluate the accuracy of the learned structure using Ar\_Precision, Ar\_Recall, and Ar\_Distance. The Ar\_Precision metric denotes the number of correctly predicted edges in the output divided by the number of true edges in a test DAG, while the Ar\_Recall metric represents the number of correctly predicted edges in the output divided by the number of predicted edges in the output of an algorithm.  The Ar\_Distance metric is the harmonic average of the Ar\_Precision and Ar\_Recall, $\rm Ar\_Distance=\sqrt{(1-Ar\_Precision)^{2}+(1-Ar\_Recall)^{2}}$, where the lower Ar\_Distance is better.

\item Efficiency. We report running time (in seconds) as the efficiency measure of different algorithms. The running time of feature selection is included in the total running time of our method.

\end{itemize}

\begin{figure*}[t]
\centering
       \begin{tabular}{ccccc}

      \subfigure{\includegraphics[width=1.55in, height=1.3in]{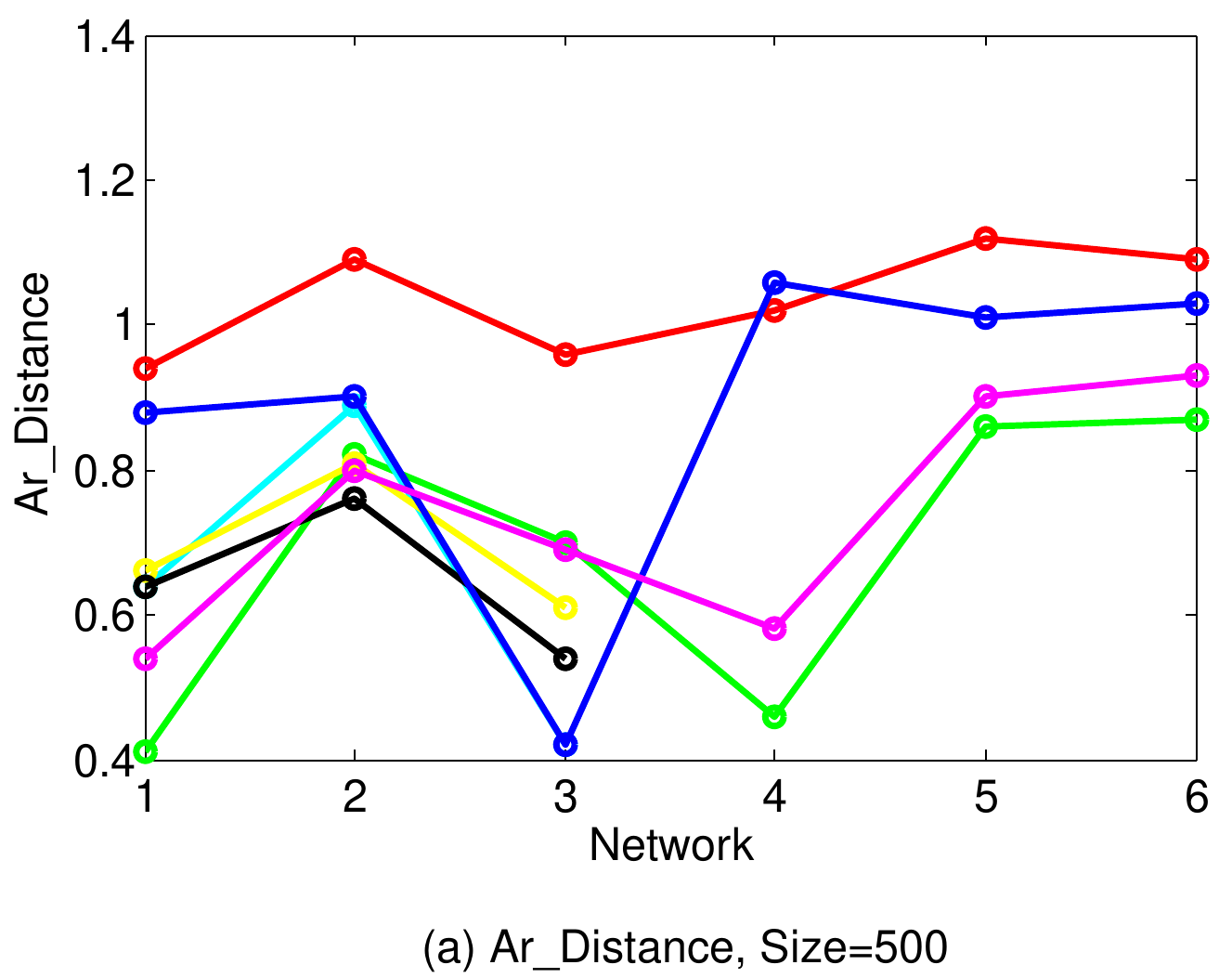}}&
      \subfigure{\includegraphics[width=1.55in, height=1.3in]{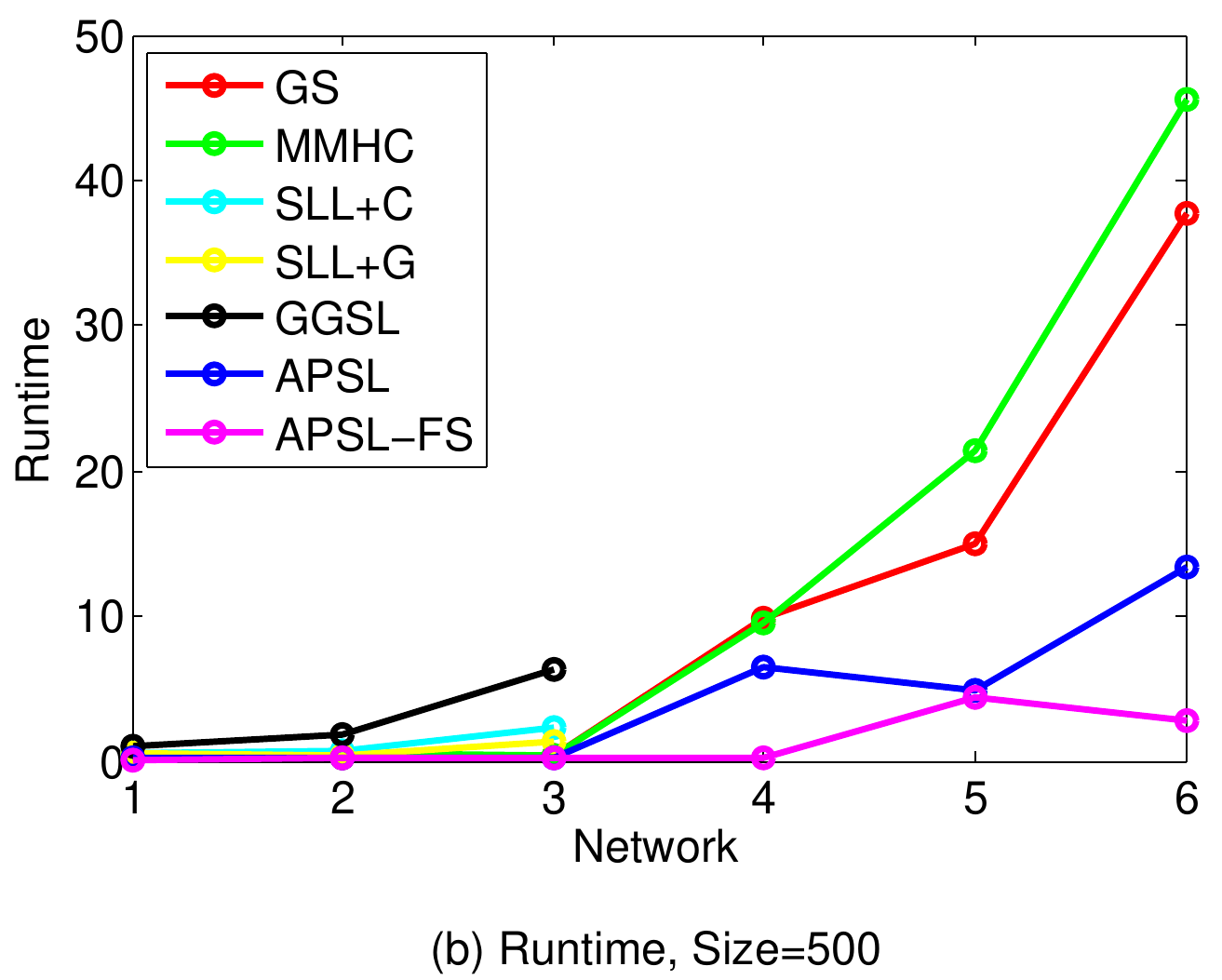}}&
      \subfigure{\includegraphics[width=1.55in, height=1.3in]{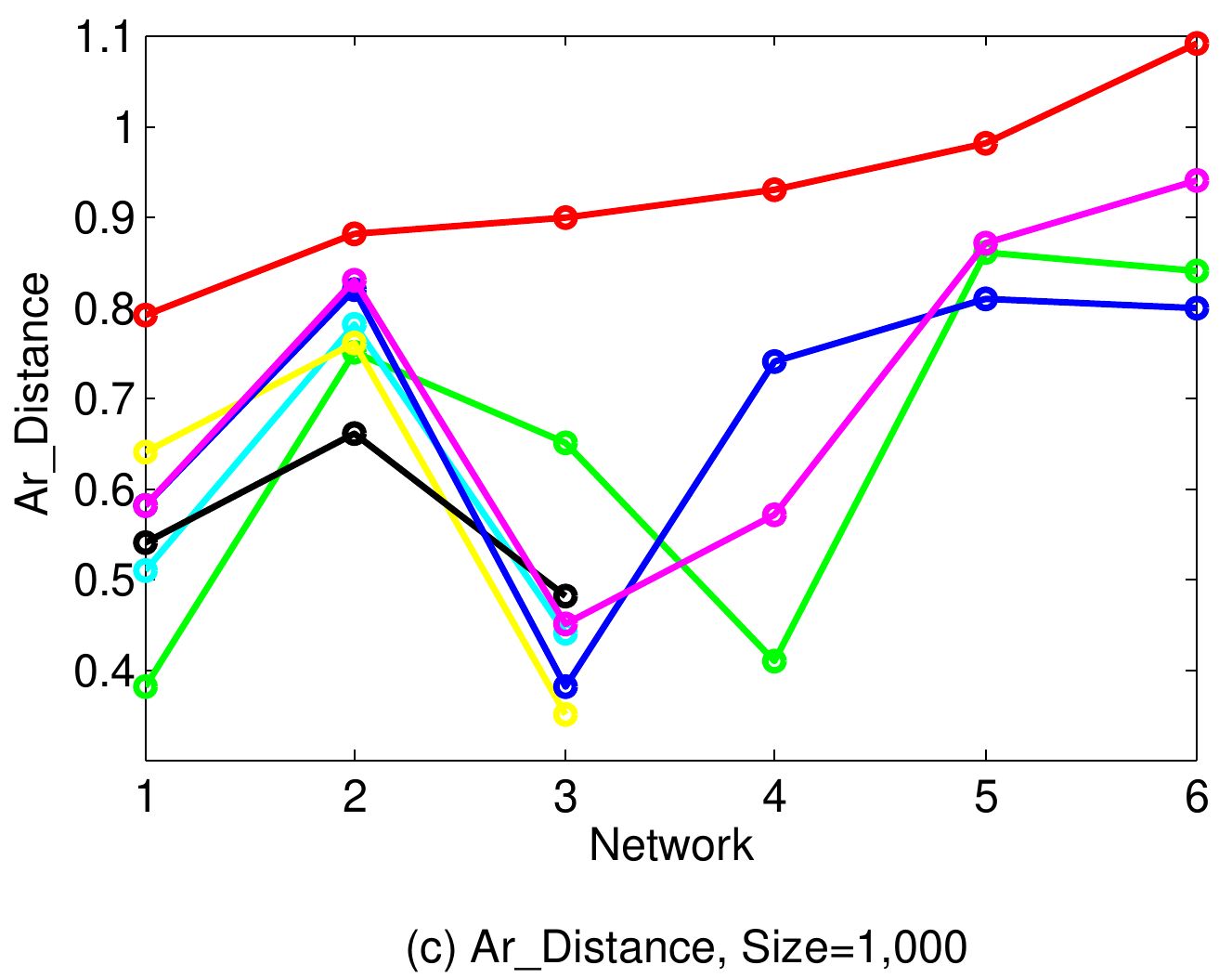}}&
      \subfigure{\includegraphics[width=1.55in, height=1.3in]{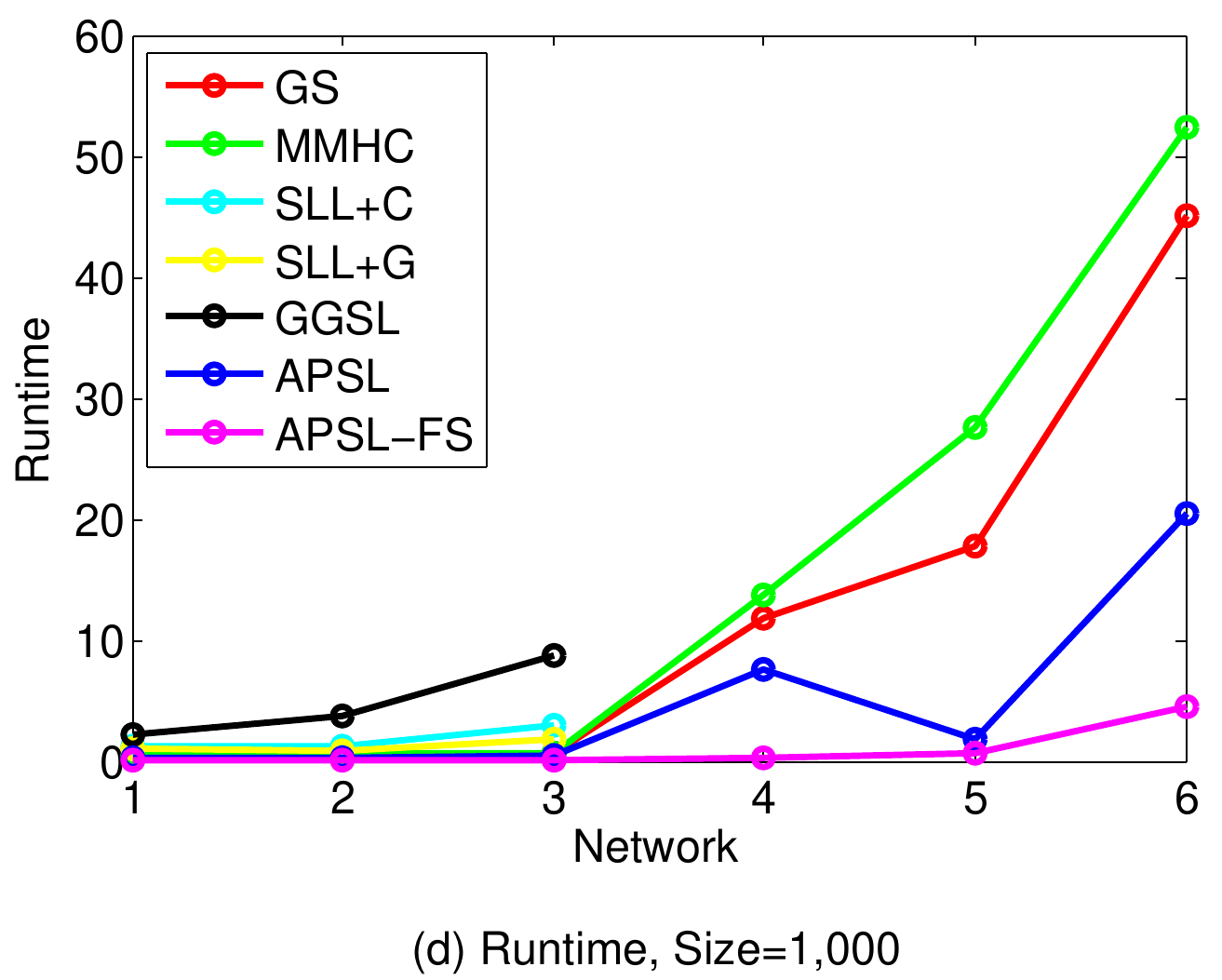}}&

       \end{tabular}
 \caption{The experimental results of learning a part of BN structures to a depth of 3 using different data sizes (the labels of the x-axis from 1 to 6 denote the BNs. 1: Child. 2: Insurance. 3: Alarm. 4: Child10. 5: Insurance10. 6: Alarm10, and all figures use the same legend).}
 \label{Figure 9}
\end{figure*}

\begin{figure*}[t]
\centering
       \begin{tabular}{ccccc}

      \subfigure{\includegraphics[width=1.55in, height=1.3in]{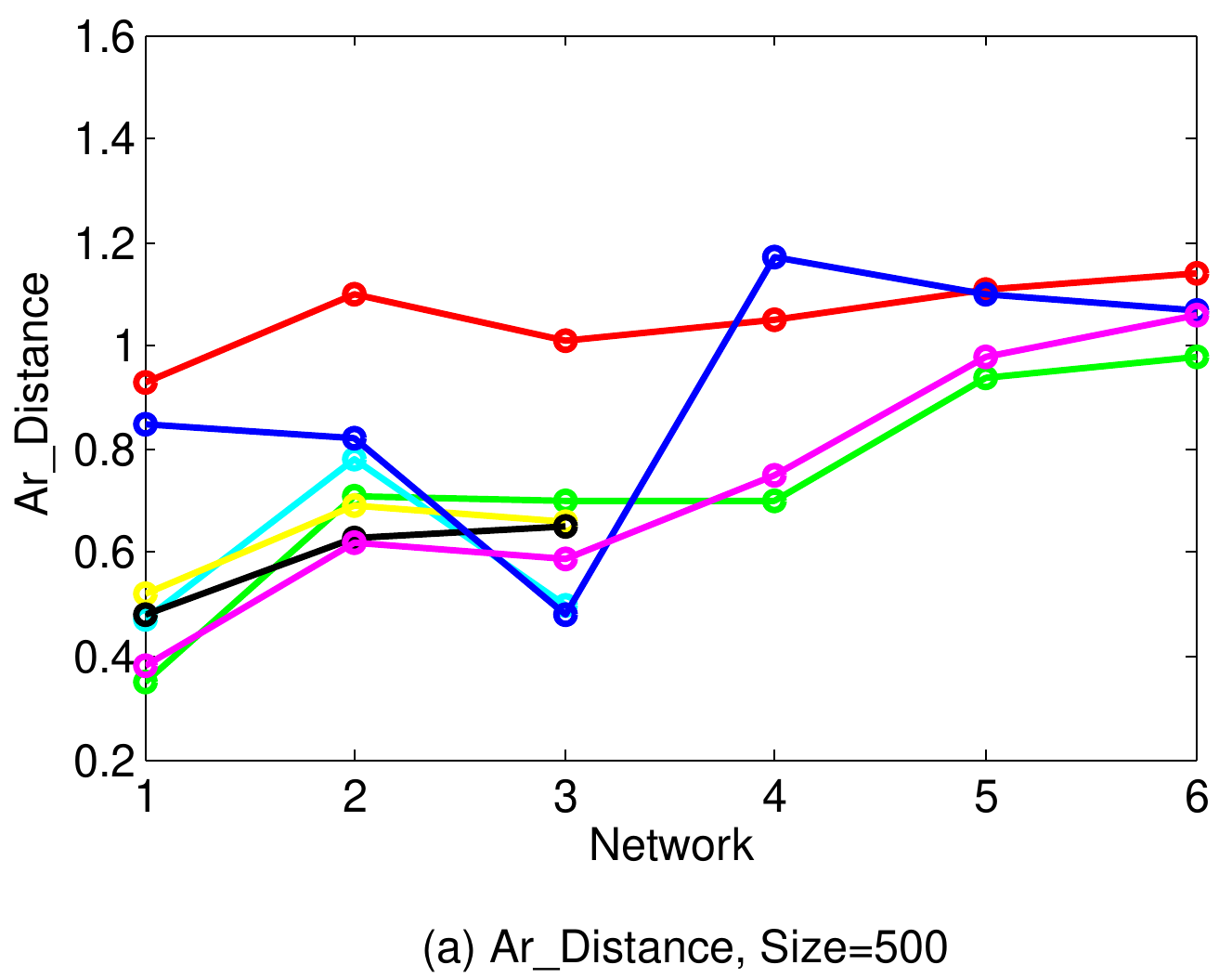}}&
      \subfigure{\includegraphics[width=1.55in, height=1.3in]{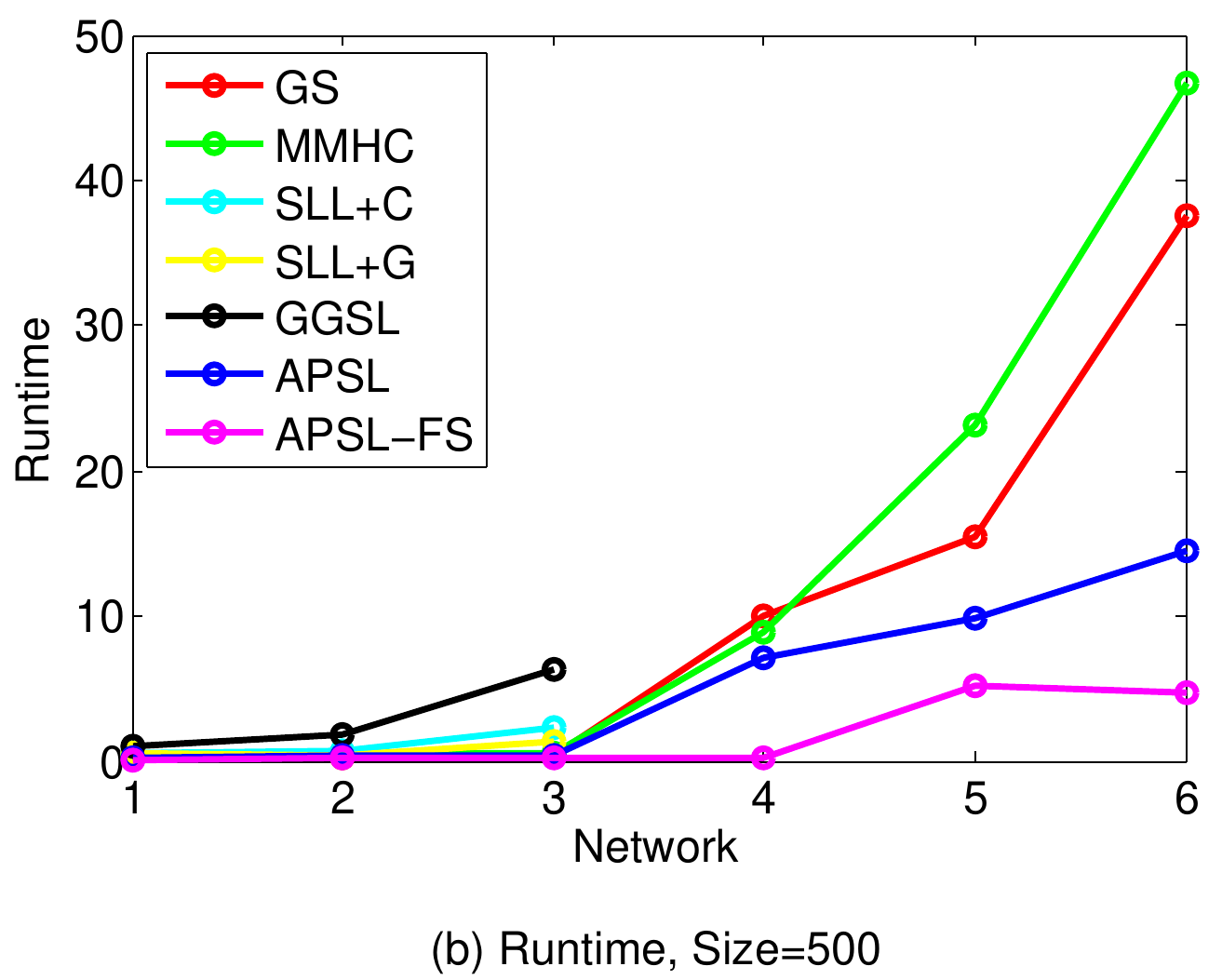}}&
      \subfigure{\includegraphics[width=1.55in, height=1.3in]{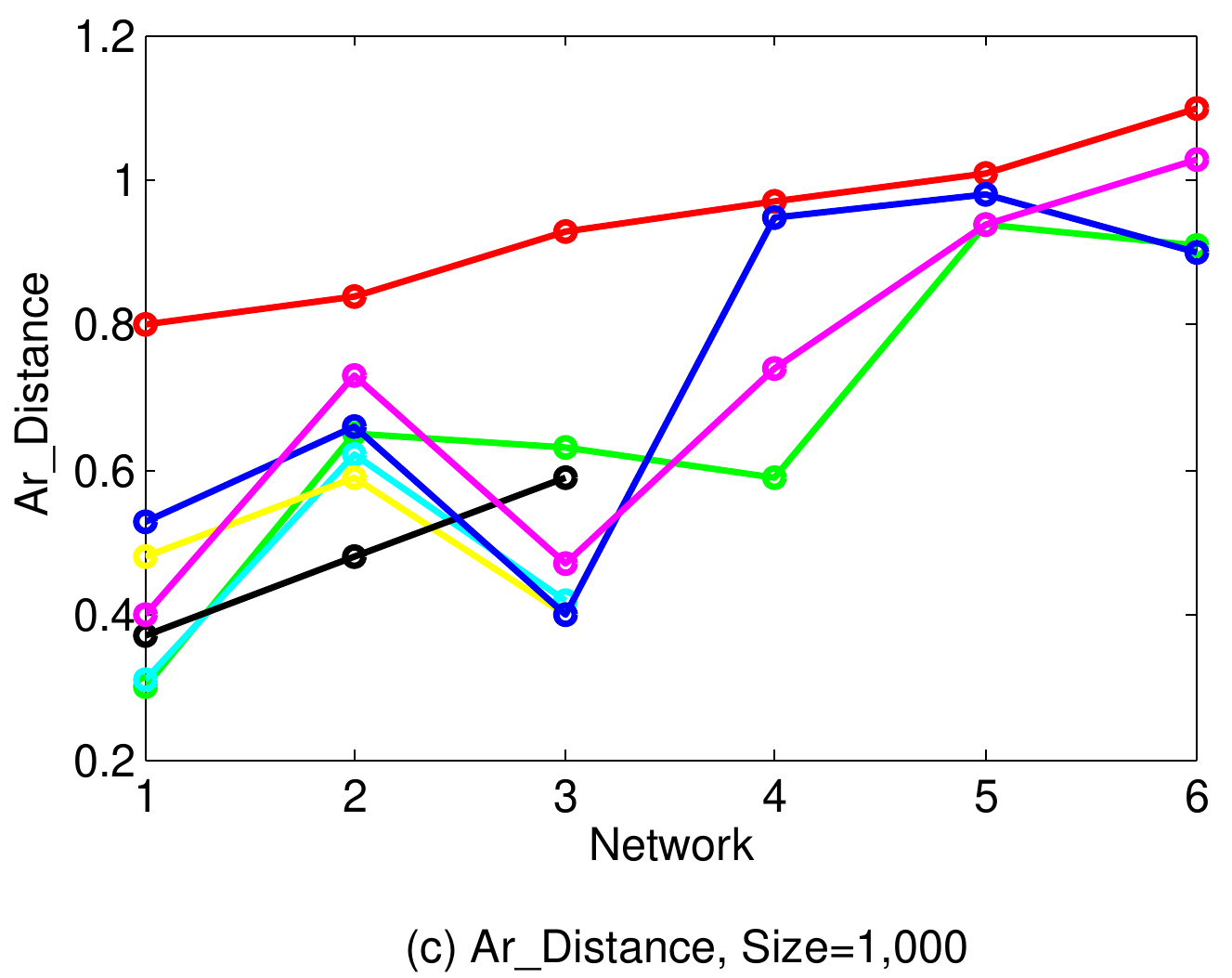}}&
      \subfigure{\includegraphics[width=1.55in, height=1.3in]{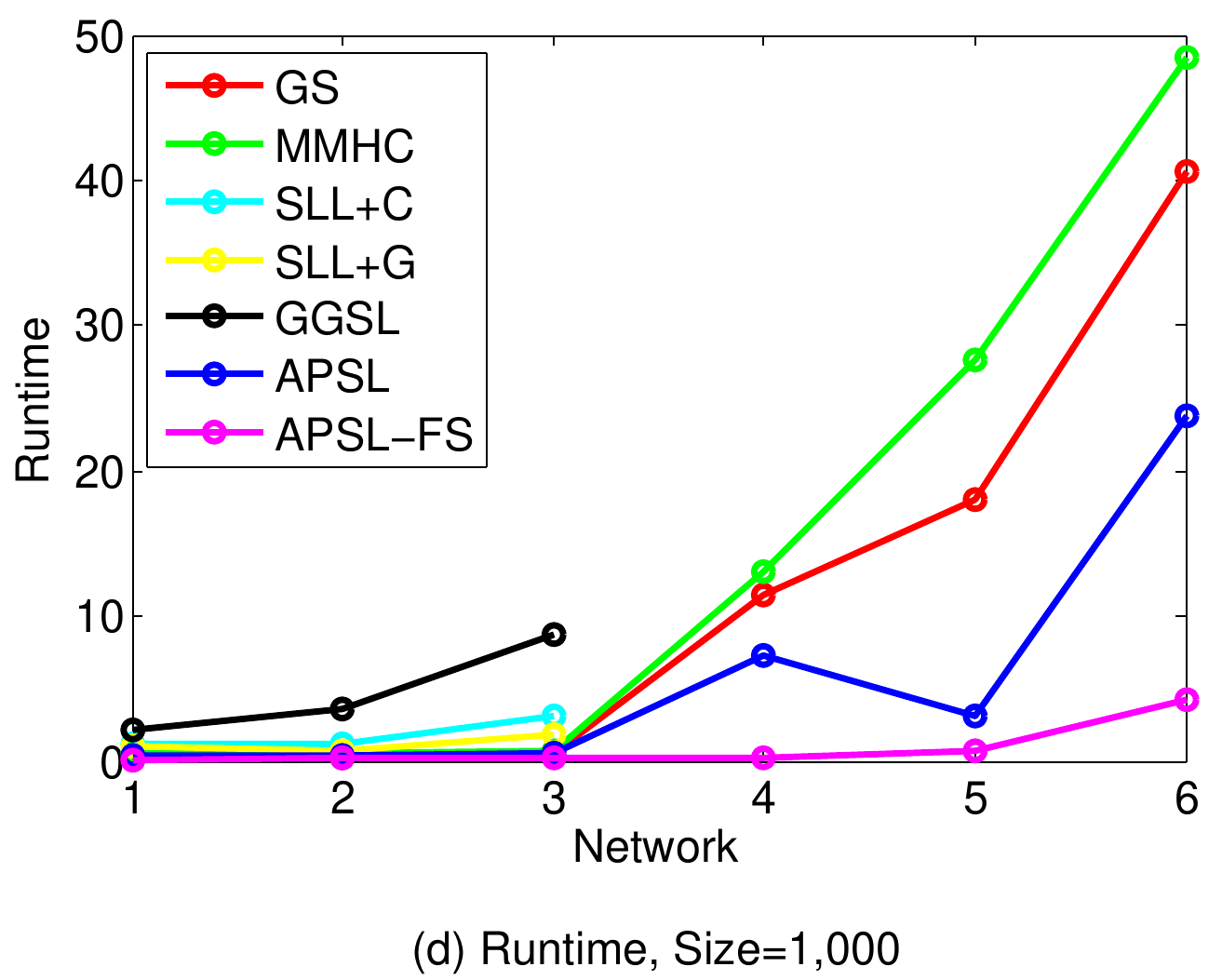}}&

       \end{tabular}
 \caption{The experimental results of learning a part of BN structures to a depth of 5 using different data sizes (the labels of the x-axis from 1 to 6 are the same as those in Fig. 10, and all figures use the same legend).}
 \label{Figure 9}
\end{figure*}

\begin{figure*}[t]
\centering
       \begin{tabular}{ccccc}

      \subfigure{\includegraphics[width=1.55in, height=1.3in]{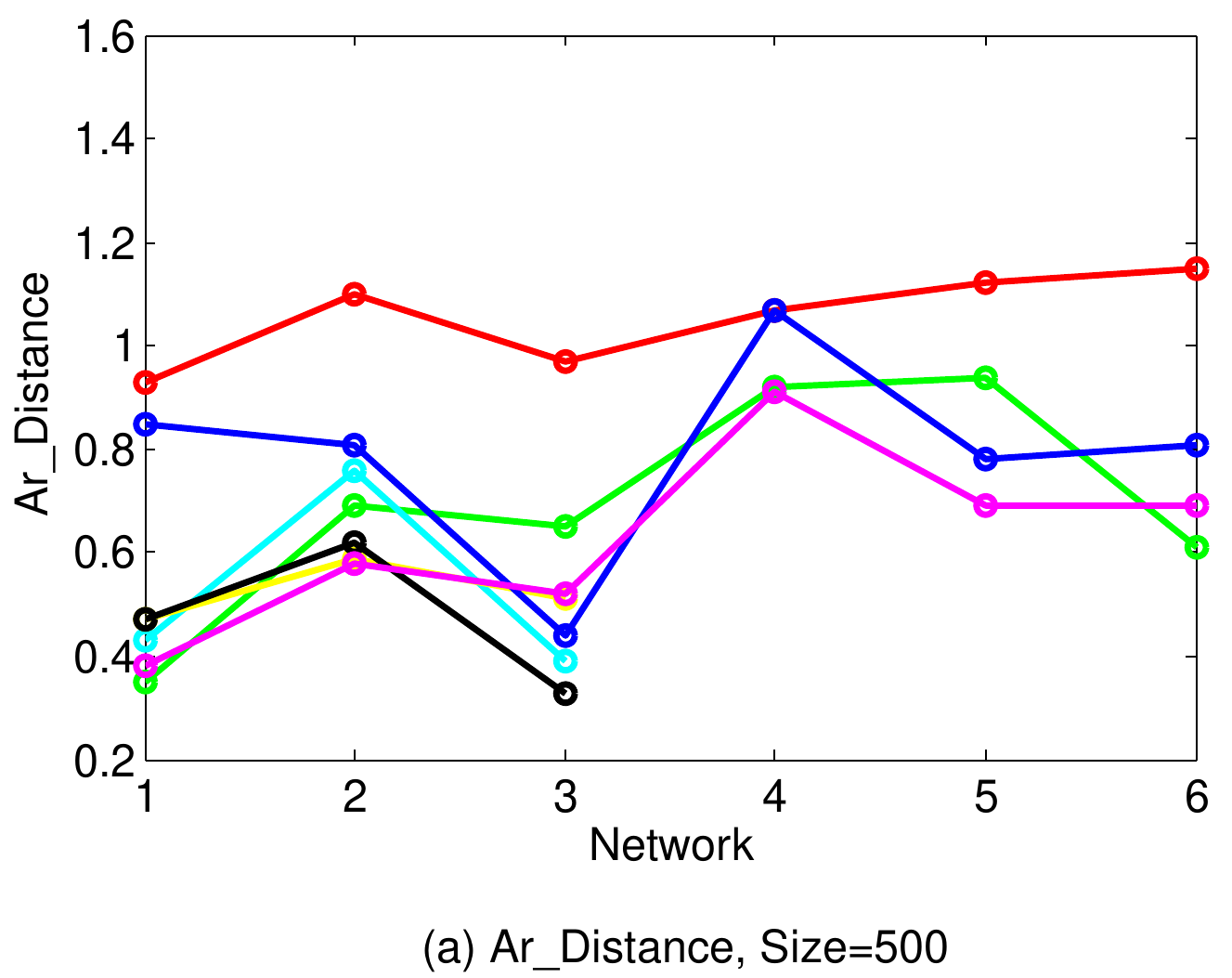}}&
      \subfigure{\includegraphics[width=1.55in, height=1.3in]{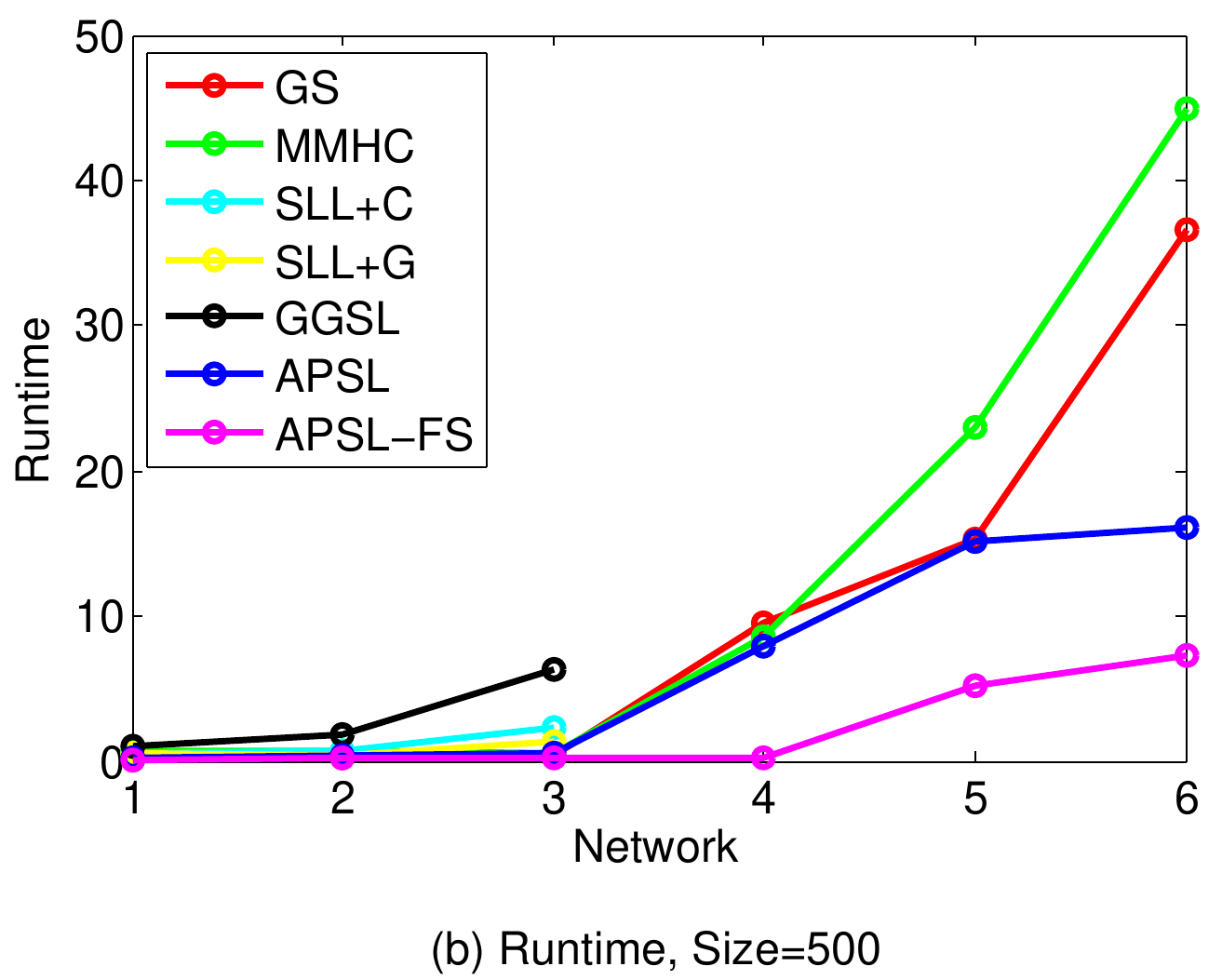}}&
      \subfigure{\includegraphics[width=1.55in, height=1.3in]{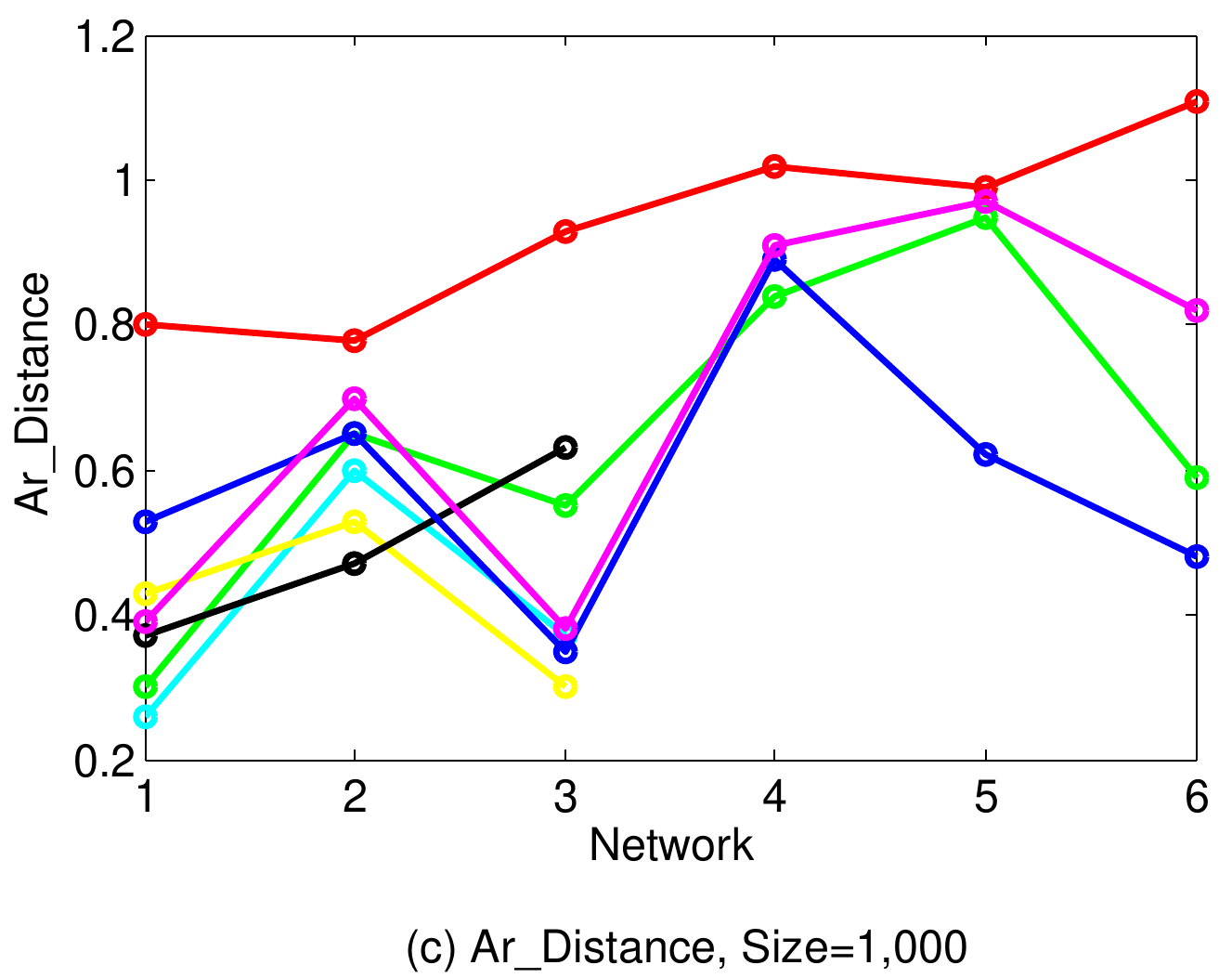}}&
      \subfigure{\includegraphics[width=1.55in, height=1.3in]{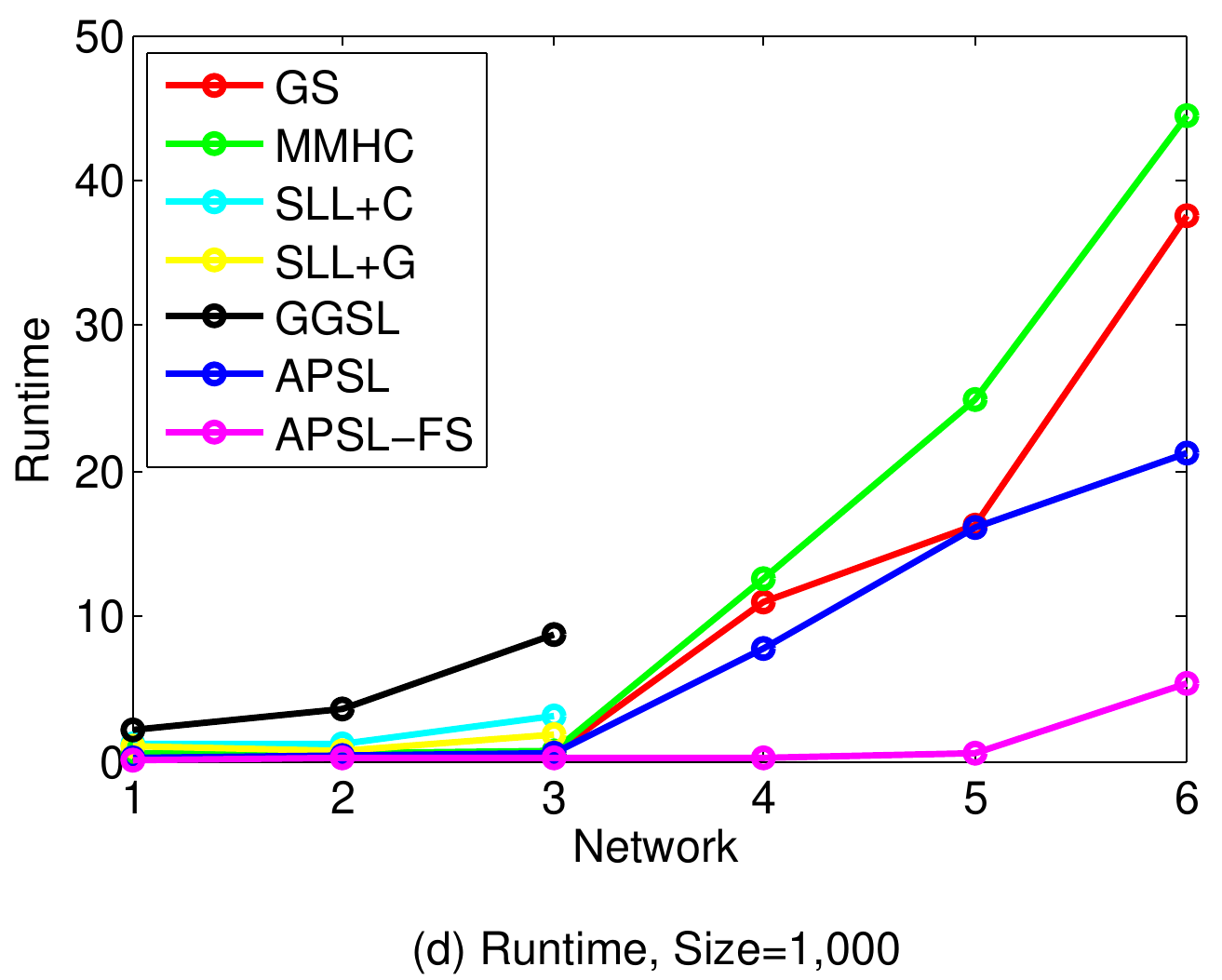}}&

       \end{tabular}
 \caption{The experimental results of learning a part of BN structures to a depth of the maximum depth using different data sizes (the labels of the x-axis from 1 to 6 are the same as those in Fig. 10, and all figures use the same legend).}
 \label{Figure 9}
\end{figure*}

\begin{figure*}[t]
\centering
       \begin{tabular}{ccccc}

      \subfigure{\includegraphics[width=2.8in, height=1.7in]{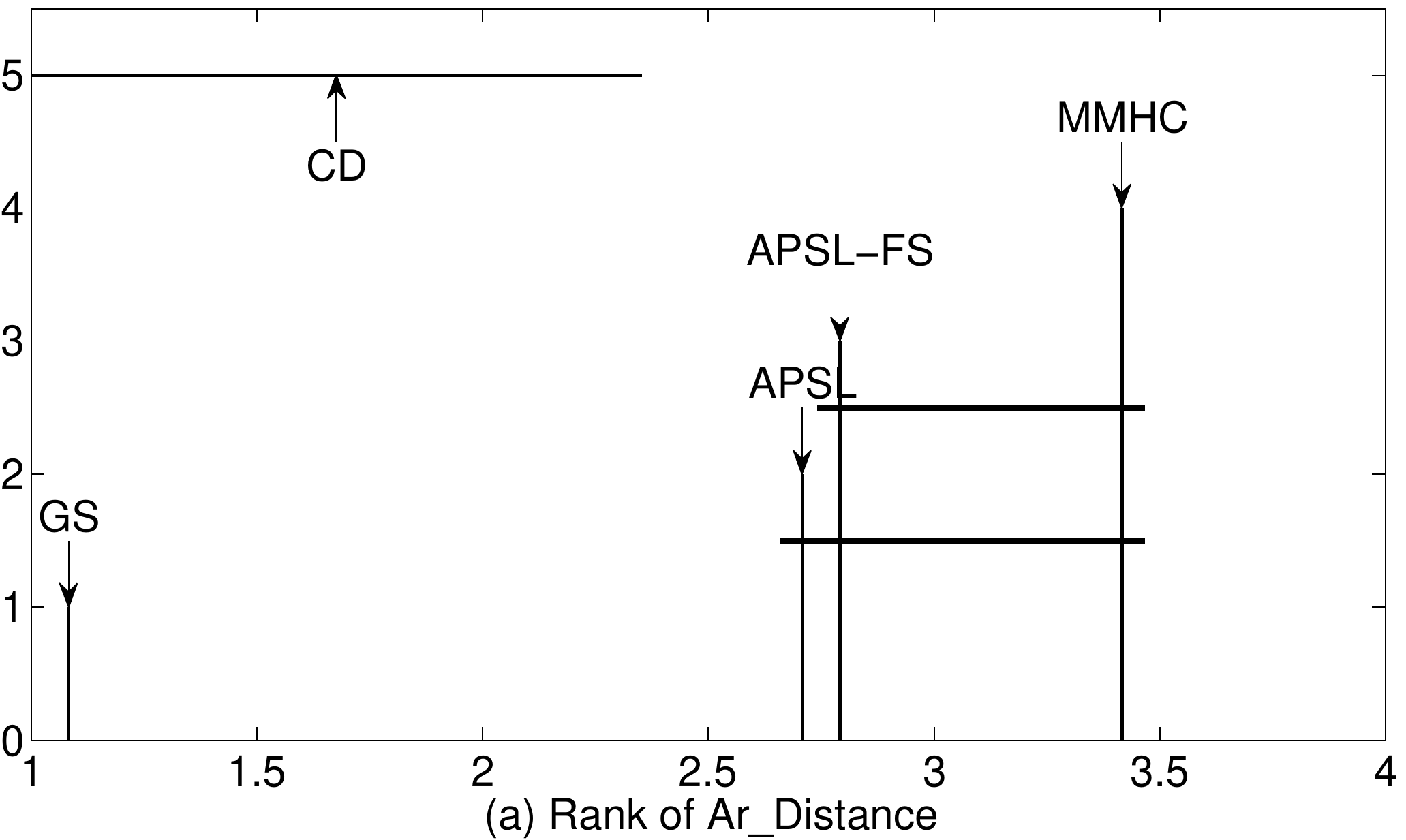}}&
      \subfigure{\includegraphics[width=2.8in, height=1.7in]{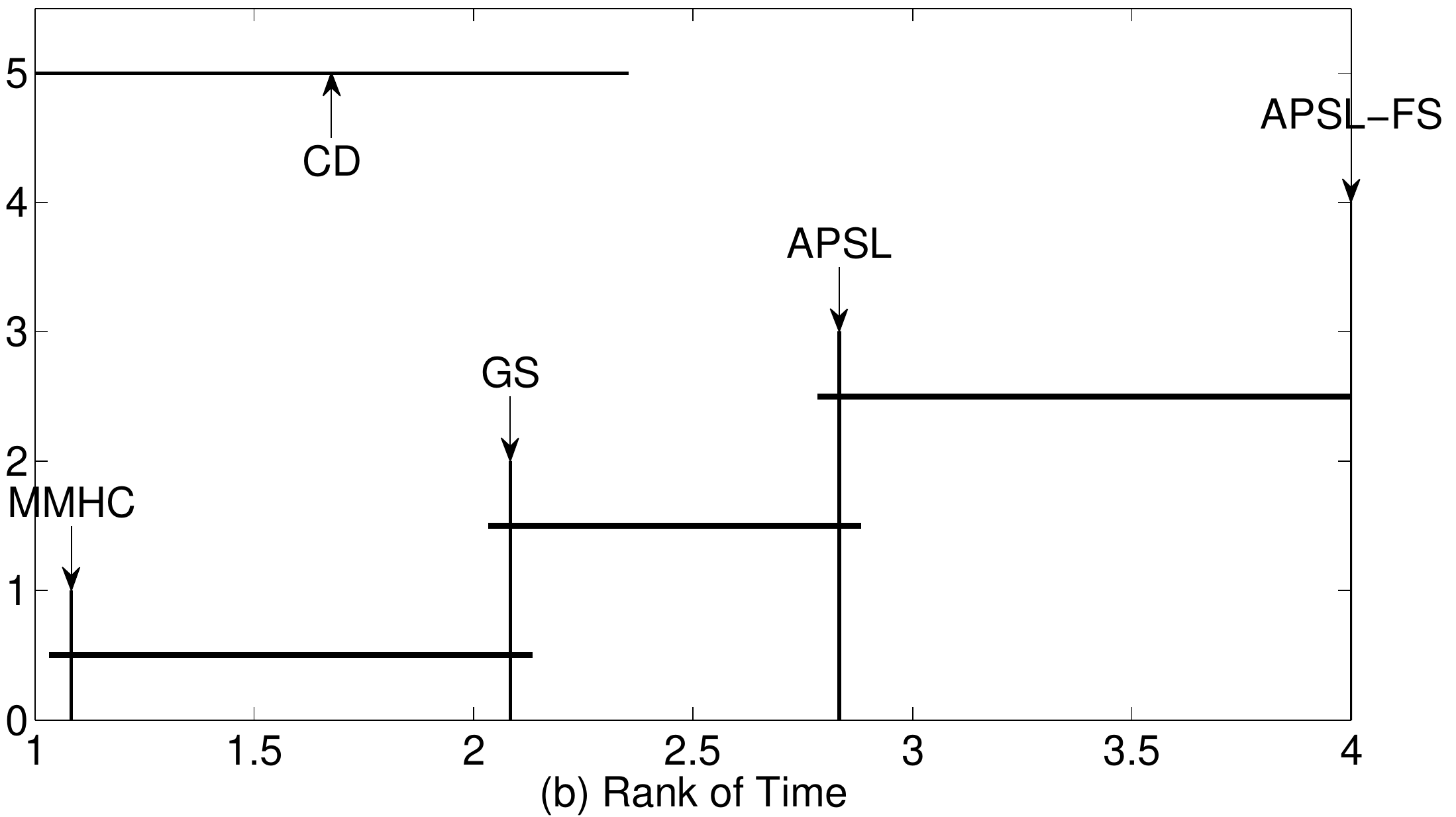}}&

       \end{tabular}
 \caption{Crucial difference diagram of the Nemenyi test of Ar\_Distance and Runtime on learning a part of BN structures (Depth=3).}
 \label{Figure 9}
\end{figure*}

\begin{figure*}[t]
\centering
       \begin{tabular}{ccccc}

      \subfigure{\includegraphics[width=2.8in, height=1.7in]{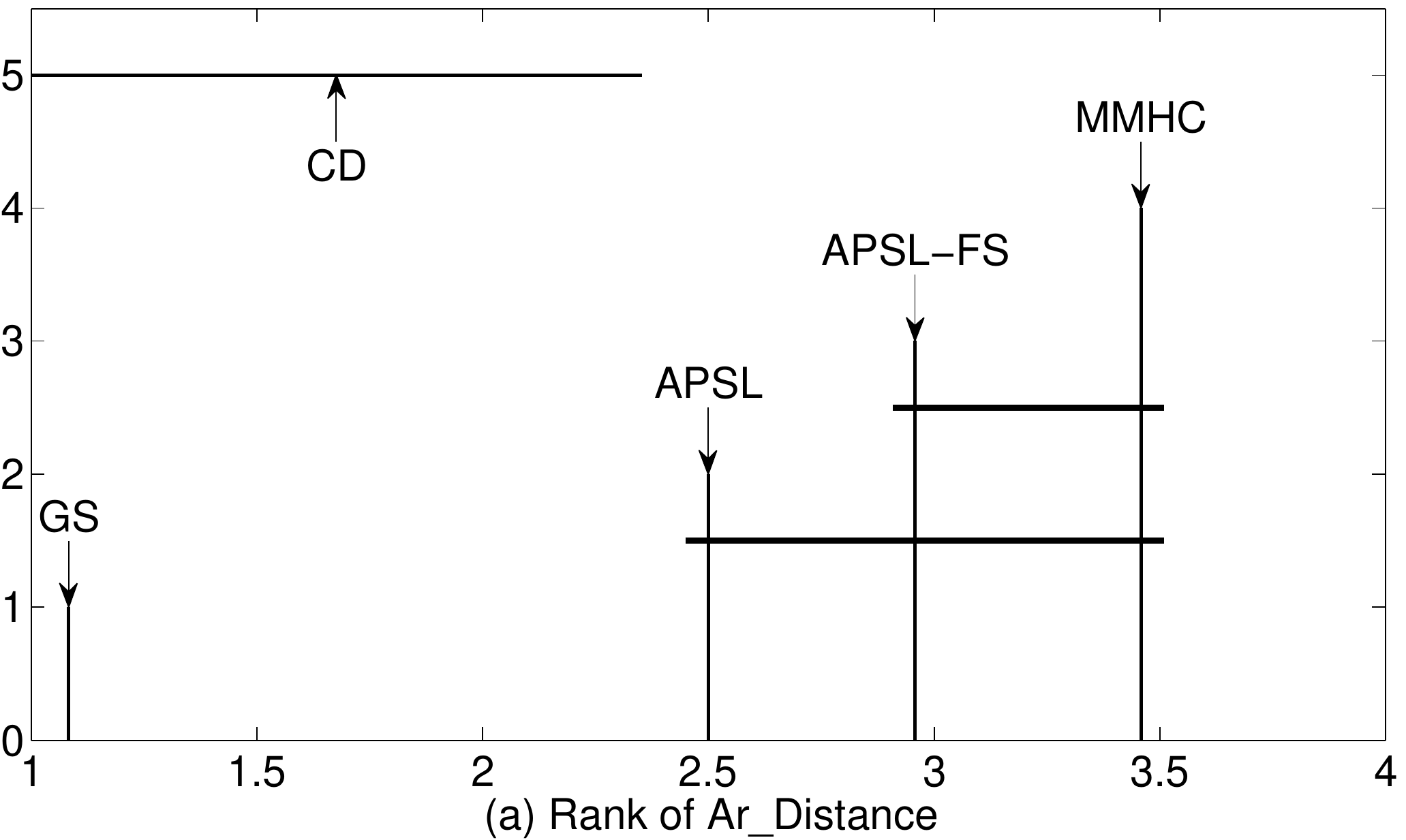}}&
      \subfigure{\includegraphics[width=2.8in, height=1.7in]{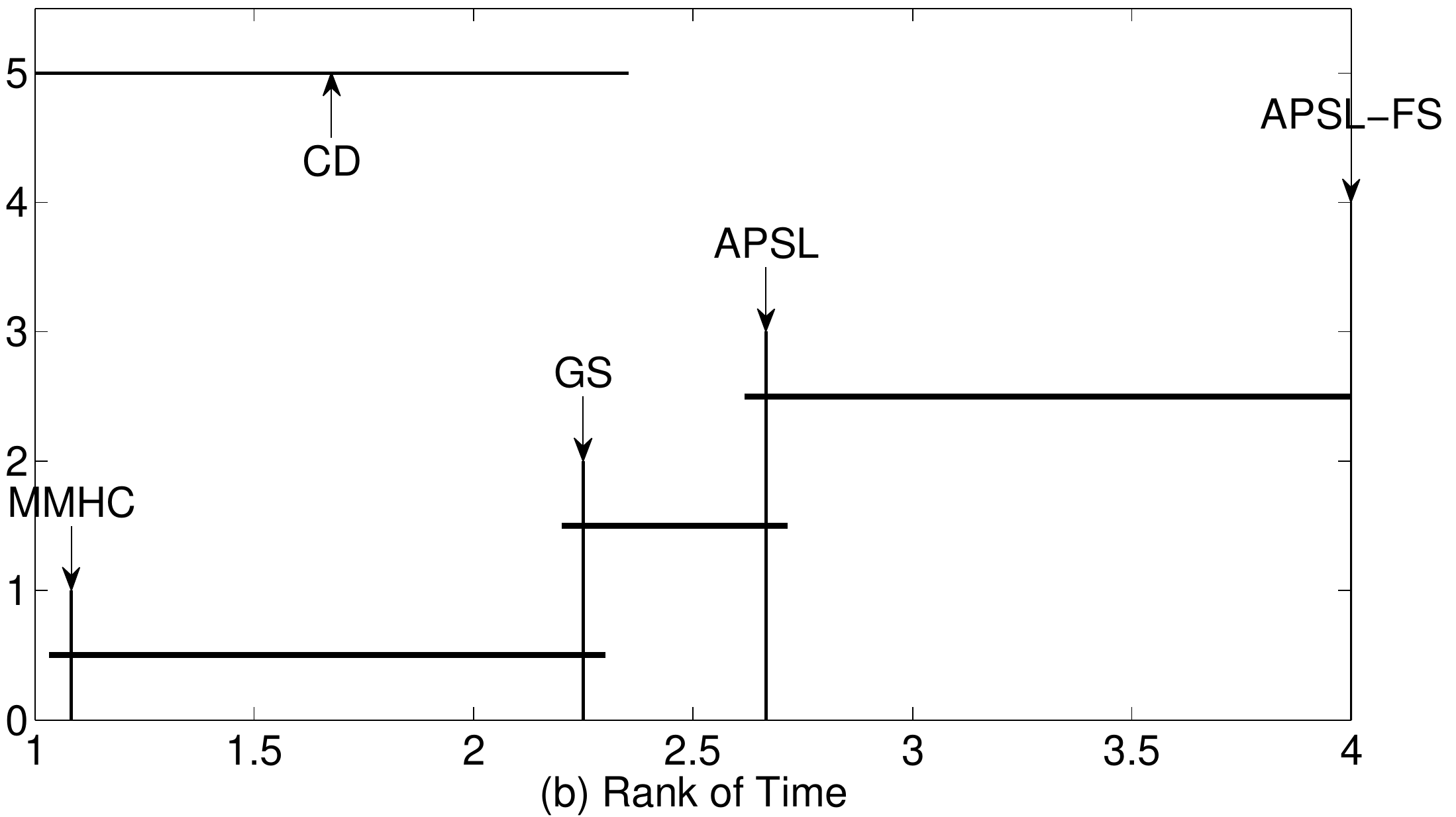}}&

       \end{tabular}
 \caption{Crucial difference diagram of the Nemenyi test of Ar\_Distance and Runtime on learning a part of BN structures (Depth=5).}
 \label{Figure 9}
\end{figure*}

\begin{figure*}[t]
\centering
       \begin{tabular}{ccccc}

      \subfigure{\includegraphics[width=2.8in, height=1.7in]{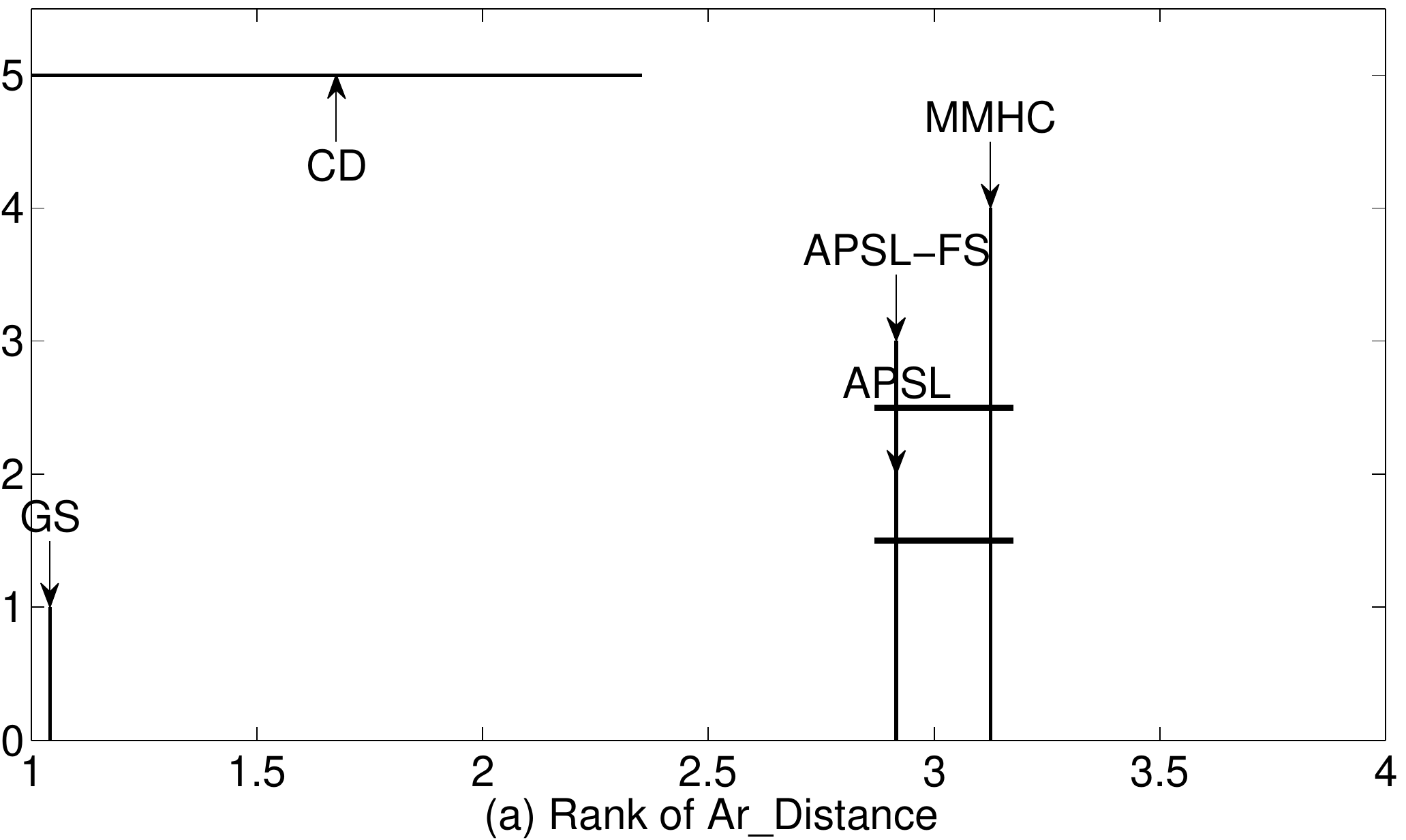}}&
      \subfigure{\includegraphics[width=2.8in, height=1.7in]{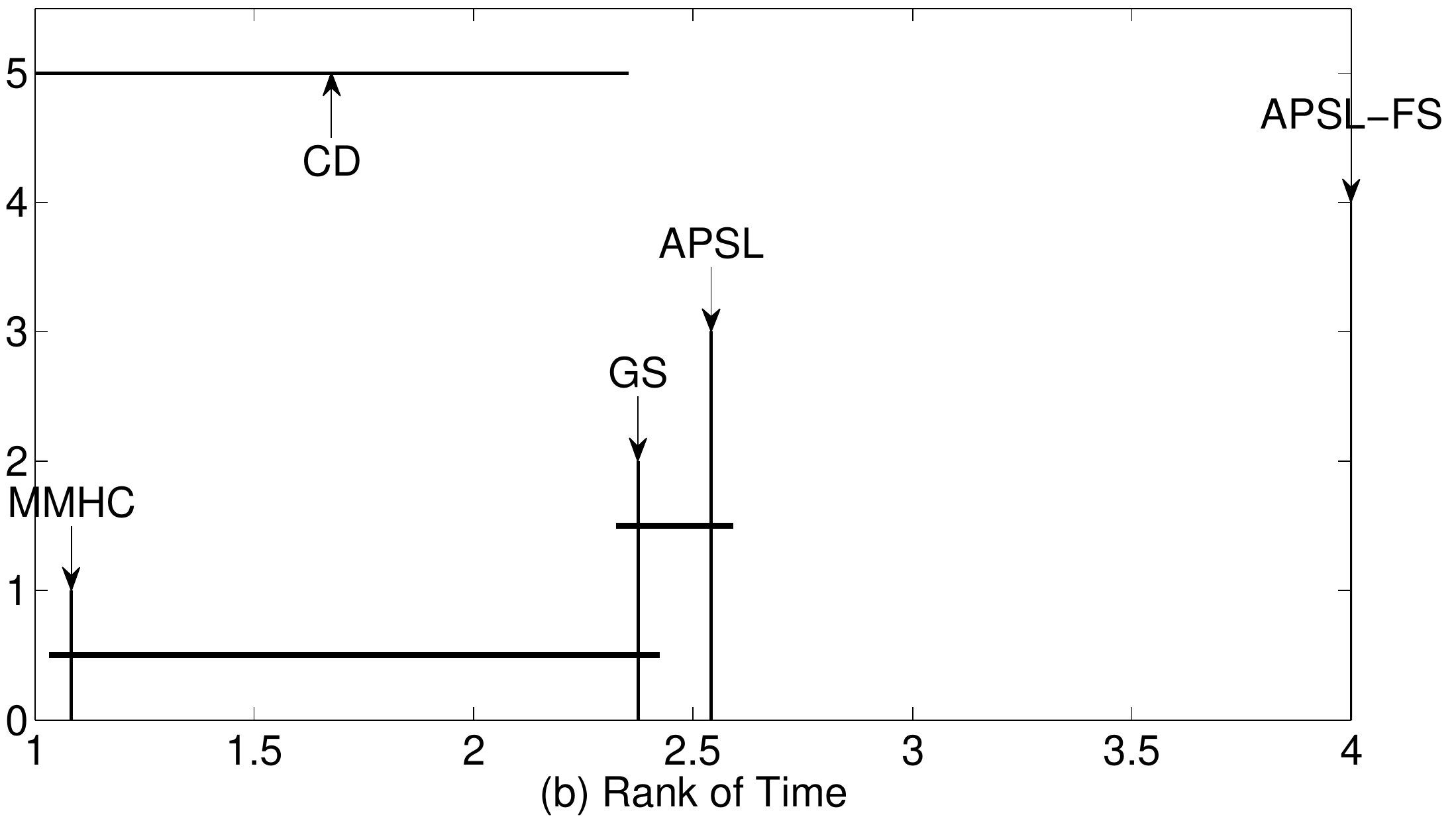}}&

       \end{tabular}
 \caption{Crucial difference diagram of the Nemenyi test of Ar\_Distance and Runtime on learning a part of BN structures (Depth=max).}
 \label{Figure 9}
\end{figure*}

\subsection{Comparison of our methods with local methods}

In this subsection, using six BNs, we compare our methods with the local methods on learning a part of a BN structure around each node to a depth of 1, Tables III summarizes the detailed results.

\emph{In efficiency}. PCD-by-PCD uses symmetry constraint to generate undirected edges, then it finds more PCs than APSL, and thus it is slower than APSL. CMB spends time tracking conditional independence changes after MB discovery, so it is inferior to APSL in efficiency. APSL-FS does not need to perform an exhaustive subset search within conditioning sets for PC discovery, then it is much faster than APSL.

\emph{In accuracy}. The symmetry constraint used by PCD-by-PCD may remove more true nodes, leading to a low accuracy of PCD-by-PCD. CMB uses entire MB set as the conditioning set for tracking conditional independence changes, so it is also inferior to APSL in accuracy. APSL-FS does not use conditioning set for independence tests, then it reduces the requirement of data samples, and more accurate than APSL on samll-sized sample data sets.

To further evaluate the accuracy and efficiency of our methods against local methods, we conduct the Friedman test at a 5\% significance level under the null hypothesis, which states that whether the accuracy and efficiency of APSL and APSL-FS and that of PCD-by-PCD and CMB have no significant difference. Both of the null hypotheses of Ar\_Distance and Runtime are rejected, the average ranks of Ar\_Distance for PCD-by-PCD, CMB, APSL, and APSL-FS are 1.54, 2.17, 3.04, and 3.25, respectively (the higher the average rank, the better the performance in accuracy), and the average ranks of Runtime for PCD-by-PCD, CMB, APSL, and APSL-FS are 1.75, 1.58, 2.83, 3.83, respectively (the higher the average rank, the better the performance in efficiency).

Then, we proceed with the Nemenyi test as a posthoc test. With the Nemenyi test, the performance of two methods is significantly different if the corresponding average ranks differ by at least the critical difference. With the Nemenyi test, both of the critical differences of Ar\_Distance and Runtime are up to 1.35. Thus, we can observe that APSL-FS is significantly more accurate than PCD-by-PCD, and APSL-FS is significantly more efficient than both of PCD-by-PCD CMB on learning a part of a BN structure to a depth of 1. We plot the crucial difference diagram of the Nemenyi test in Fig. 8.

\begin{table}[!htbp]
\caption{Five nodes with the largest PC set on each BN}{
\footnotesize
\begin{center}
\begin{tabular}{llccc}
\toprule

Network     & Selected five nodes\\

\midrule

Child       & 2,     6,     7,     9,     11    \\
Insurance   & 2,     3,     4,     5,     8     \\
Alarm       & 13,    14,    21,    22,    29    \\
Child10     & 12,    52,    92,    132,   172   \\
Insurance10 & 164,   166,   191,   193,   245   \\
Alarm10     & 13,    23,    66,    103,   140   \\

\bottomrule
\end{tabular}
\end{center}}
\end{table}

\subsection{Comparison of our methods with global methods}

In this subsection, we compare our methods with the global methods on learning a part of a BN structure to a depth of 3, 5, and the maximum depth, respectively. 


In Fig. 9-11, we plot the results of Ar\_Distance and Runtime of APSL and APSL-FS against global methods on learning part of BN structures around the five nodes with the largest PC set on each BN to a depth of 3, 5, and the maximum, respectively. The selected five nodes of each BN are shown in Table IV. Since SLL+C, SLL+G, and GGSL cannot generate any results on $Child10$, $Insurance10$, and $Alarm10$ due to memory limitation, we only plot the results of them on $Child$, $Insurance$, and $Alarm$.

\emph{In efficiency}. When learning a part of BN structures to depths of 3 and 5, since APSL and APSL-FS do not need to find the entire structures, they are faster than the global BN structure learning algorithms. When learning a part of BN structures to a depth of the maximum depth, both of our methods and global methods need to find the entire structure. However, \emph{1)} Although GS uses GSMB, an efficient MB discovery algorithm without searching for conditioning set, to find MB of each node, it still takes extra time to search for conditioning set during V-structure discovery. So GS is slightly inferior to APSL in efficiency. \emph{2)} May be using conditional independence tests is faster than using score functions for edge orientations, then MMHC is slower than APSL. \emph{3)} As for SLL+C, SLL+G, and GGSL, the score-based MB/PC methods used by them need to learn a local BN structure involving all nodes selected currently at each iteration, so they are time-consuming on small-sized BNs, and infeasible on large-sized BNs. \emph{4)} Clearly, APSL-FS is more efficient than APSL.

\emph{In accuracy}. When learning a part of BN structures to depths of 3 and 5, since global methods consider the global information of the structures, the accuracy of our methods is lower that of global methods except for GS. Because the GSMB (used by GS) require a large number of data samples, and its heuristic function also leads to a low MB discovery accuracy. When learning a part of BN structures to a depth of the maximum depth, \emph{1)} since the same reason of GS when learning to a depth of 3 and 5, GS is inferior to our methods in accuracy. \emph{2)} MMHC uses score functions for edge orientations, it can also remove false edges in the learned skeleton, while APSL can only orient edges in the learned skeleton using conditional independence tests, then MMHC is more accurate than APSL. \emph{3)} As for SLL+C, SLL+G, and GGSL, since they involve all nodes selected currently at each iteration, they are slightly more accurate than other methods on small-sized BNs, but cannot generate any results on large-sized BNs. \emph{4)} Similarly, APSL-FS is more accurate than APSL.

To further evaluate the accuracy and efficiency of our methods against global methods, we conduct the Friedman test at a 5\% significance level under the null hypothesis. Since SLL+C, SLL+G, and GGSL fail on the large-sized BN data sets, we do not compare our methods with them using the Friedman test.

\emph{1) Depth=3}. Both of the null hypotheses of Ar\_Distance and Runtime are rejected, the average ranks of Ar\_Distance for GS, MMHC, APSL, and APSL-FS are 1.08, 3.42, 2.71, and 2.79, respectively, and the average ranks of Runtime for GS, MMHC, APSL, and APSL-FS are 2.08, 1.08, 2.83, and 4.00, respectively. Then, With the Nemenyi test, both of the critical differences of Ar\_Distance and Runtime are up to 1.35. Thus, we can observe that APSL and APSL-FS are significantly more accurate than GS and significantly more efficient than MMHC, and APSL-FS is significantly more efficient than GS on learning a part of a BN structure to a depth of 3. We plot the crucial difference diagram of the Nemenyi test in Fig. 12.

\emph{2) Depth=5}. Similar to the results in Depth=3, the average ranks of Ar\_Distance for GS, MMHC, APSL, and APSL-FS are 1.08, 3.46, 2.50, and 2.96, respectively, and the average ranks of Runtime for GS, MMHC, APSL, and APSL-FS are 2.25, 1.08, 2.67, and 4.00, respectively. With the critical differences of Ar\_Distance and Runtime are up to 1.35, we can observe that APSL and APSL-FS are significantly more accurate than GS and significantly more efficient than MMHC, and APSL-FS is significantly more efficient than GS on learning a part of a BN structure to a depth of 5. We plot the crucial difference diagram of the Nemenyi test in Fig. 13.

\emph{3) Depth=max}. Similarly, the average ranks of Ar\_Distance for GS, MMHC, APSL, and APSL-FS are 1.04, 3.13, 2.92, and 2.92, respectively, and the average ranks of Runtime for GS, MMHC, APSL, and APSL-FS are 2.38, 1.08, 2.54, and 4.00, respectively. With the critical differences of Ar\_Distance and Runtime are up to 1.35, we can observe that APSL and APSL-FS are significantly more accurate than GS and significantly more efficient than MMHC, and APSL-FS is significantly more efficient than GS on learning a part of a BN structure to a depth of the maximum. We plot the crucial difference diagram of the Nemenyi test in Fig. 14.

\section{Conclusion}

In this paper,
we present a new concept of Expand-Backtracking to describe the learning process of the exsiting local BN structure learning algorithms, and analyze the missing V-structures in Expand-Backtracking.
Then we propose an efficient and accurate any part of BN structure learning algorithm, APSL. APSL learns a part of a BN structure around any one node to any depth, and tackles missing V-structures in Expand-Backtracking by finding both of collider V-structures and non-collider V-structures in MBs at each iteration.
In addition, we design an any part of BN structure learning algorithm using feature selection, APSL-FS, to improve the efficiency of APSL by finding PC without searching for conditioning sets.

The extensive experimental results have shown that our algorithms achieve higher efficiency and better accuracy than state-of-the-art local BN structure learning algorithms on learning any part of a BN structure to a depth of 1, and achieve higher efficiency than state-of-the-art global BN structure learning algorithms on learning any part of a BN structure to a depth of 3, 5, and the maximum depth.

Future research direction could focus on using mutual information-based feature selection methods for V-structure discovery without searching for conditioning sets, because performing an exhaustive subset search within PC for finding V-structures is time-consuming.


\bibliographystyle{IEEEtran}
\bibliography{References}

\end{document}